\documentclass{article} %
\usepackage{iclr2027_conference,times}
\usepackage{natbib}

\usepackage[accsupp]{axessibility}  %

\usepackage{amsmath,amsfonts,bm}

\def\eqref#1{equation~\ref{#1}}

\def\1{\bm{1}}

\DeclareMathAlphabet{\mathsfit}{\encodingdefault}{\sfdefault}{m}{sl}
\SetMathAlphabet{\mathsfit}{bold}{\encodingdefault}{\sfdefault}{bx}{n}

\PassOptionsToPackage{nameinlink}{cleveref}
\usepackage[dvipsnames,table]{xcolor}
\definecolor{cvprblue}{rgb}{0.21,0.49,0.74}
\definecolor{darkred}{rgb}{0.70,0.13,0.13} %
\usepackage{hyperref}
\hypersetup{
    breaklinks=true,
    colorlinks=true,
    citecolor=cvprblue,
    linkcolor=darkred,
    urlcolor=cvprblue
}

\usepackage{graphicx}
\usepackage{booktabs}  %
\usepackage[utf8]{inputenc} %
\usepackage[T1]{fontenc}
\usepackage{url}            %
\usepackage{amsfonts}       %
\usepackage{nicefrac}       %
\usepackage{microtype}      %
\usepackage{gensymb}
\usepackage[inline]{enumitem}
\usepackage{tabularx}
\usepackage{caption}
\usepackage{subcaption}
\usepackage{multirow}
\usepackage{svg}
\usepackage{cuted}
\usepackage{tabu}
\usepackage{pifont}  %
\usepackage{mwe}
\usepackage{bm}  %
\usepackage{amsmath}
\usepackage{textcomp}
\usepackage{makecell}  %
\usepackage{grffile} %
\usepackage{rotating}  %
\usepackage{graphics} %
\usepackage{adjustbox}
\usepackage[capitalize]{cleveref}
\usepackage{wrapfig}
\usepackage{soul}
\usepackage{svg}
\usepackage{url}

\usepackage{eccvabbrv}

\usepackage{colortbl}

\definecolor{neurips}{rgb}{0.12,0.49,0.85}

\usepackage{siunitx}  %
\DeclareSIUnit{\million}{\text{M}}
\newcommand{\comment}[1]{}

\renewcommand{\paragraph}[1]{\vskip4pt \noindent\textbf{#1}}

\definecolor{tabfirst}{rgb}{1, 0.7, 0.7} %
\definecolor{tabsecond}{rgb}{1, 0.85, 0.7} %
\definecolor{tabthird}{rgb}{1, 1, 0.7} %

\definecolor{rankfirst}{HTML}{7CB696}
\definecolor{ranksecond}{HTML}{BEDBBD}
\definecolor{rankthird}{HTML}{E8F1E4}
\definecolor{groupgray}{HTML}{EEEEEE}
\newcommand{\fst}{\cellcolor{rankfirst}\bfseries}
\newcommand{\snd}{\cellcolor{ranksecond}}
\newcommand{\trd}{\cellcolor{rankthird}}

\newcommand{\norm}[1]{\left\lVert#1\right\rVert}

\makeatletter
\DeclareRobustCommand\onedot{\futurelet\@let@token\@onedot}
\def\@onedot{\ifx\@let@token.\else.\null\fi\xspace}
\def\eg{\emph{e.g}\onedot} 
\def\ie{\emph{i.e}\onedot}

\makeatletter
\g@addto@macro\@maketitle{\begin{center}
  \begin{minipage}[c]{0.43\linewidth}  %
    \centering
    {\footnotesize\strut\textbf{SfM on ETH3D:} accuracy vs.\ throughput}\\[1pt]
    \includegraphics[width=\linewidth]{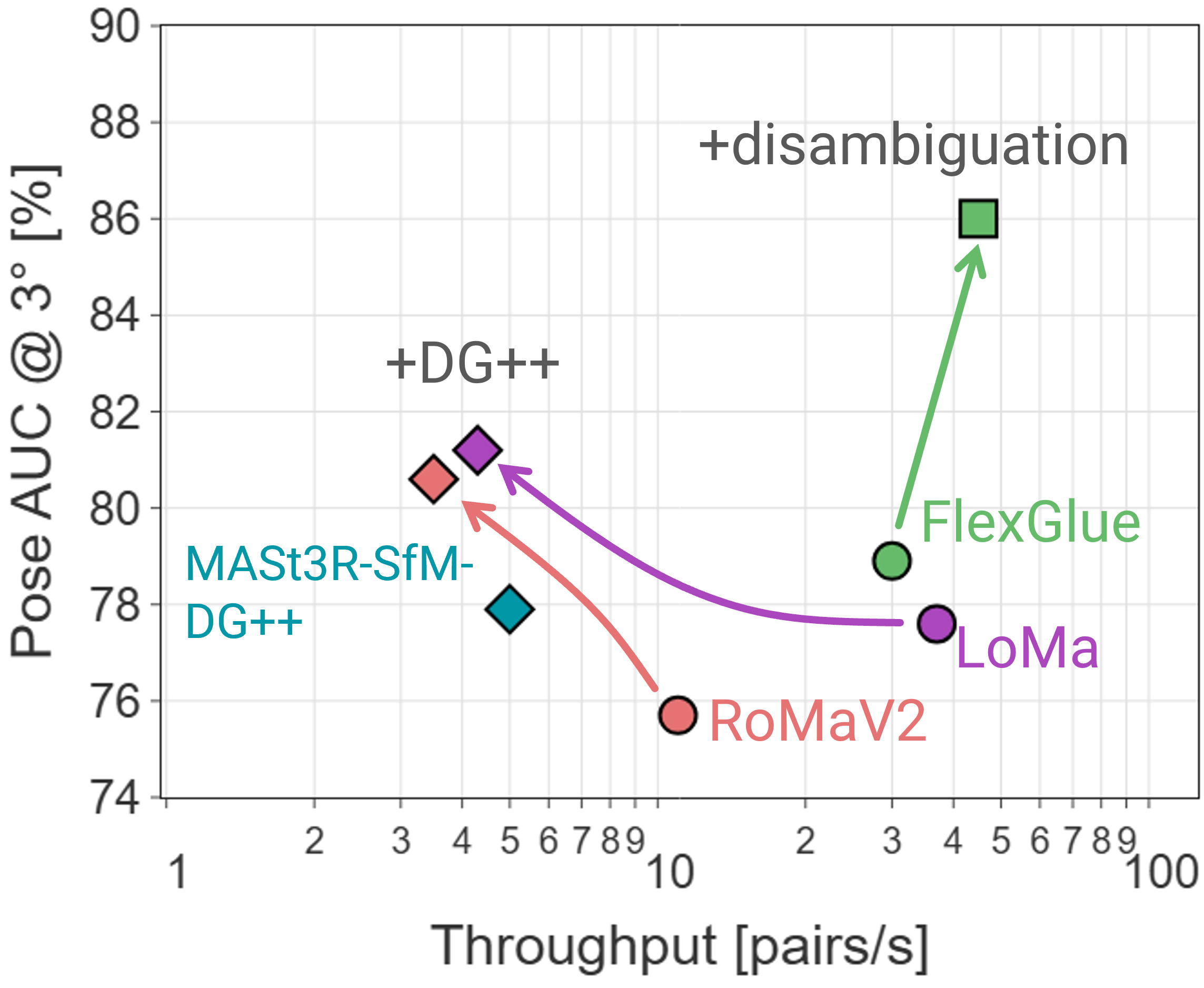}
  \end{minipage}\hfill
  \begin{minipage}[c]{0.56\linewidth}
    \centering
    \includegraphics[width=\linewidth]{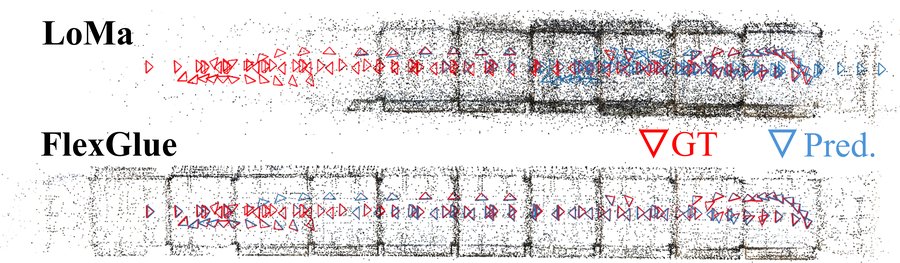}\\[2pt]  %
    \definecolor{teasergreen}{RGB}{40,150,80}
    \definecolor{teaserred}{RGB}{205,70,70}%

    \begin{minipage}[t]{\dimexpr(\linewidth-2mm)/2-0.5pt\relax}
      \centering
      \includegraphics[width=\linewidth]{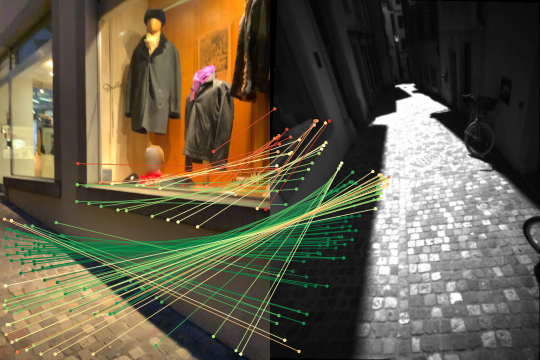}
      {\footnotesize\strut\textcolor{teasergreen}{\textbf{Covisible:} 122 matches}}\\[1pt]  %

    \end{minipage}\hfill
    \begin{minipage}[t]{\dimexpr(\linewidth-2mm)/2-0.5pt\relax}
      \centering

      \includegraphics[width=\linewidth]{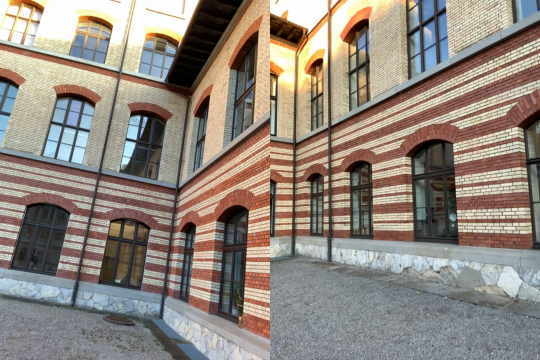}
      {\footnotesize\strut\textcolor{teaserred}{\textbf{Doppelganger:} 98\%}}\\[1pt]
    \end{minipage}
  \end{minipage}
  \captionof{figure}{%
  \textbf{To match or not to match.}
  Deep matchers are trained only on covisible pairs and thus return many confident matches between distinct but similar-looking structures, which breaks Structure-from-Motion  (top right).
  FlexGlue tracks 2D query points across an image pair with subpixel accuracy and jointly predicts whether the two images are covisible, rejecting such doppelgangers with high confidence (bottom right).
  For 3D applications, FlexGlue is more robust than state-of-the-art matchers and disambiguates more accurately than the dedicated classifier DG++~\citep{xiangli2025doppelgangers++} (left).
  We show that disambiguation even improves throughput by stopping early on non-covisible pairs.
  }
  \label{fig:teaser-new}
\end{center}
}
\makeatother

\title{FlexGlue: Sparse-to-Dense Feature Matching and Visual Disambiguation}

\iclrfinalcopy
\author{Philipp Lindenberger$^{1}$ \And Paul-Edouard Sarlin$^{2}$ \And Ondrej Miksik$^{3}$ \And Marc Pollefeys$^{1,3}$
\AND
{\normalfont%
$^{1}$ETH Zurich
\hspace{.4in}%
$^{2}$Google
\hspace{.4in}%
$^{3}$Microsoft Spatial AI Lab
}
}

\begin{document}

\maketitle
\lhead{} %

\begin{abstract}
To estimate correspondences across images, existing approaches match either sparse interest points, which is highly efficient but limited by repeatability, or dense pixel grids, which is robust but expensive.
Furthermore, these approaches are trained with only matching pairs and thus often fail to reject unrelated pairs that exhibit visual symmetries, which is equally important in applications like Structure-from-Motion (SfM).
We bridge these gaps by introducing FlexGlue, a neural network for asymmetric sparse-to-dense matching and disambiguation.
Given two images and coordinates of points in the first image, FlexGlue estimates the coordinates of the corresponding points in the second image while simultaneously rejecting the pair if non-matching.
By searching for a limited set of correspondences among dense pixels, our approach is both fast and effective at both matching and disambiguation:
it combines the sub-pixel accuracy of dense matching with the speed of sparse matching, while rejecting incorrect pairs with high precision.
Experiments on multiple datasets show that FlexGlue pushes the Pareto frontier in relative pose estimation and significantly outperforms existing approaches for SfM and visual localization.
FlexGlue is publicly available at \href{https://github.com/cvg/flexglue}{\texttt{github.com/cvg/flexglue}}.

\end{abstract}

\begin{figure*}[t]
  \centering
  \includegraphics[width=0.97\linewidth]{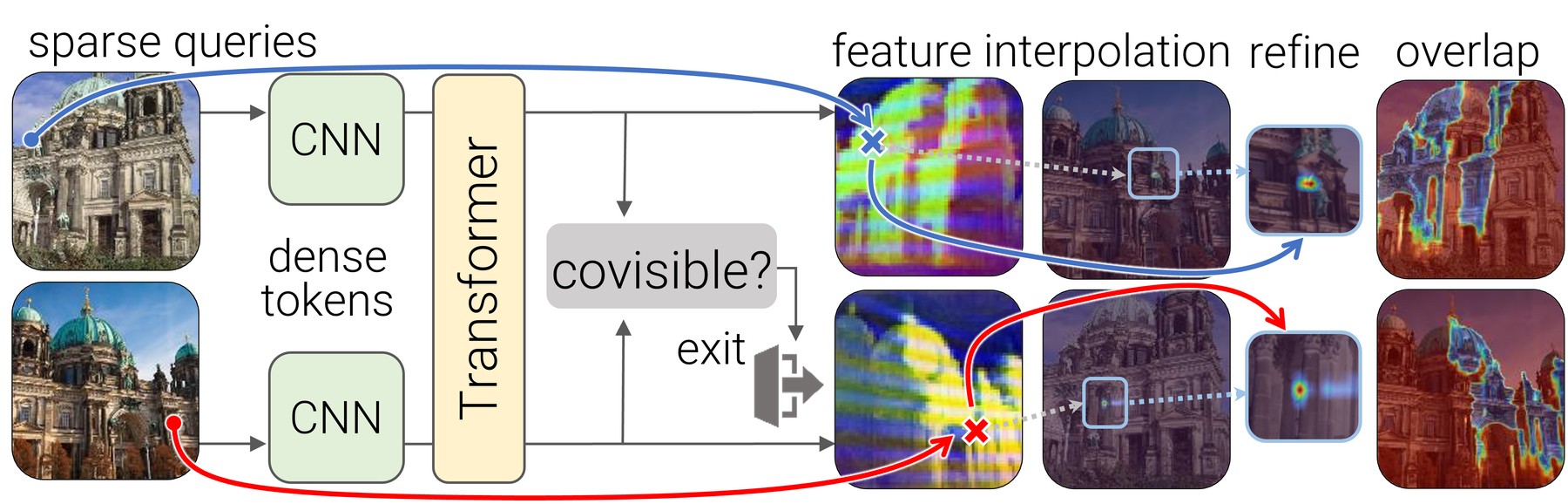}  %
  \caption{%
  \textbf{Architecture of FlexGlue.}
  Each of the two images is encoded by a CNN backbone.
  The resulting dense tokens are fused by a Transformer that outputs context-augmented tokens.
  These tokens are decoded into overlap heatmaps and classify whether the image pair is unmatchable, in which case the inference stops.
  Otherwise, a query point in one image is turned into a sub-pixel coordinate in the other by interpolation of its feature, dense correlation, and match refinement.
  }
  \label{fig:pipeline}
\end{figure*}

\section{Introduction}

Image matching estimates correspondences across images and is a fundamental building block of 3D vision, from Structure-from-Motion (SfM) to visual localization.
Most geometric applications only require correspondences for a sparse set of distinctive interest points~\citep{lowe2004distinctive, yi2016lift, superpoint}, which both simplifies the problem and reduces its cost.
These points are detected independently in each image and matched by a nearest-neighbor search of their descriptors or by deep networks that learn priors on motion and scene structure~\citep{sarlin2020superglue, lindenberger2023lightglue, nordstrom2026loma}.
Matching is thus fundamentally bounded by the detector: a 3D point can only be matched if it is detected in both views, and any localization noise of the detections propagates directly into the correspondences.

Dense matchers~\citep{DKM,edstedt2023roma,leroy2024grounding,edstedt2026romav2} lift this bound by matching all pixels directly, and are remarkably robust to large viewpoint and appearance changes while achieving subpixel accuracy.
This comes at a high cost: they estimate and refine a correspondence for every pixel, which restricts them to low resolutions and makes training expensive, even though downstream applications typically use only a sparse subset.
Multi-view applications such as SfM further require discrete tracks, forcing pipelines to snap dense correspondences to detected keypoints with additional heuristics~\citep{sun2021loftr, DF-SFM}.
Semi-dense matchers~\citep{sun2021loftr, li2025edm} reduce the cost by matching on a coarse grid, but anchor correspondences to grid cells and refine all of them, even when only a sparse subset is needed.
Existing matchers thus face a clear trade-off between the robustness of dense and the efficiency of sparse matching.

Beyond this trade-off, learned sparse and dense matchers share a global blind spot: they are trained exclusively on covisible image pairs.
Real-world applications such as SfM and SLAM, however, must reject pairs that depict different places, and matchers that confidently match visually similar structures, \eg, symmetric facades, cause catastrophic scene collapses, where classical methods like SIFT~\citep{lowe2004distinctive} often still perform best.
Recent works therefore add bespoke models that predict covisibility after matching~\citep{cai2023doppelgangers, xiangli2025doppelgangers++}, which introduces architectural redundancy and cost, although local matching and global disambiguation are intrinsically the same task: finding overlapping context.

To bridge these gaps, we introduce FlexGlue (\cref{fig:teaser-new}), which matches sparse queries against dense context: given an arbitrary query point in one image, it predicts its subpixel correspondence in the other.
A lightweight Transformer~\citep{vaswani2017attention} aggregates context over a dense token grid of both images, and a cheap readout and refinement head extracts correspondences only at the queries (\cref{fig:pipeline}).
Correspondences can thus land on any pixel, removing the dependence on repeatable detections while retaining the efficiency of sparse matching.
As a result, FlexGlue is substantially cheaper to train and run than dense matchers, while matching or even surpassing them on relative pose estimation, SfM, and visual localization.

Crucially, FlexGlue is trained jointly to match covisible pairs and to reject non-covisible ones, exploiting the synergy between matching and visual disambiguation.
Since its dense context aggregation already builds a global understanding of the image pair, a lightweight covisibility head suffices to outperform bespoke disambiguation models.
As training data with symmetries is scarce, we further show that disambiguation can be learned label-free from flip augmentations.
Detecting negative pairs in early layers moreover accelerates matching, in contrast to prior efficient matchers~\citep{lindenberger2023lightglue}, which stop early on easy covisible pairs but run the full network on unmatchable ones.

 In summary, our main contributions are as follows:
 \begin{enumerate}[label=(\roman*), leftmargin=*, topsep=2pt, itemsep=1pt, parsep=0pt, partopsep=0pt]
     \item \textbf{Efficient sparse-to-dense matching:} a simple architecture that combines the efficiency of sparse matchers with the robustness and accuracy of dense matchers, achieving state-of-the-art accuracy on some two-view estimation and indoor visual localization benchmarks.
     \item \textbf{Joint matching and disambiguation:} a joint formulation that outperforms dedicated disambiguation models at $4\times$ higher throughput and stops early on unmatchable pairs. It significantly improves the robustness of SfM on ETH3D and the CVPR 2025 Image Matching Challenge.
     \item \textbf{Indoor doppelganger benchmark:} the first indoor and cross-device benchmark for visual disambiguation, which we found is a crucial blindspot of existing methods, built from LaMAR~\citep{sarlin2022lamar} with hard-mined negatives. Our method improves by a large margin there.
 \end{enumerate}

\section{Related Work}
\paragraph{Sparse feature matching} detects keypoints independently in each image~\citep{lowe2004distinctive, yi2016lift, superpoint, tyszkiewicz2020disk, revaud2019r2d2, shenoi2026raco, Zhao2023ALIKED} and matches them by a nearest-neighbor search of their descriptors~\citep{superpoint, HardNet, tyszkiewicz2020disk, Zhao2023ALIKED} or with deep networks that predict a partial assignment~\citep{sarlin2020superglue, lindenberger2023lightglue, nordstrom2026loma}.
These approaches are efficient and naturally yield the discrete tracks used in multi-view geometry, but since detection is independent per image, their accuracy is upper-bounded by the repeatability and localization noise of the detector.
Moreover, their attention aggregates context only at the keypoint locations, even when guided by foundation-model features as in OmniGlue~\citep{jiang2024omniglue}, which limits the global understanding of the scene, in particular in textureless or repetitive regions.
In contrast, FlexGlue aggregates context over a dense grid independently of the keypoints and uses them only as queries, retaining the efficiency and the discrete tracks of sparse matchers.

\paragraph{Dense and semi-dense matching} circumvents the repeatability bottleneck by estimating correspondences directly on image grids~\citep{DKM, edstedt2023roma, edstedt2026romav2, zhang2025ufm}.
Dense matchers achieve impressive robustness and accuracy by refining a correspondence for every pixel in a coarse-to-fine scheme, whose cost grows with the image resolution.
MASt3R~\citep{leroy2024grounding} additionally regresses dense 3D pointmaps.
Semi-dense matchers~\citep{sun2021loftr, wang2024eloftr, li2025edm} reduce this cost with a coarse assignment on a low-resolution grid, but anchor correspondences to its cells, which complicates the extraction of discrete multi-view tracks~\citep{DF-SFM}.
Our sparse-to-dense formulation aggregates context only on a coarse grid and reads out subpixel correspondences at sparse queries, whose cost scales linearly with their number. Querying keypoints in both images further yields discrete tracks.

\paragraph{Sparse-to-dense matching} bypasses keypoint detection in the target image by matching sparse queries against its dense features.
S2DNet~\citep{germain2020s2dnet} casts this as a dense classification over CNN hypercolumns, and CAPS~\citep{wang2020learning} uses it to learn descriptors from camera poses only.
Both rely on local receptive fields and lack the global context needed to resolve ambiguities, while COTR~\citep{jiang2021cotr} has global context but refines each query by slow recursive zoom-ins.
RCM~\citep{lu2024raising} follows the design of semi-dense matchers but replaces the coarse grid of one view, chosen per pair, by sparse keypoints, which limits its context in that view to the keypoints.
Differently, FlexGlue aggregates context over the full dense grids of both images.
\paragraph{Visual disambiguation} is critical to reject non-covisible pairs and visual symmetries, which otherwise cause catastrophic failures in 3D reconstruction.
Pipelines shortlist pairs with global image retrieval~\citep{arandjelovic2016netvlad, izquierdo2024optimal, Berton_2025_CVPR} and verify them by matching and view-graph reasoning~\citep{heinly2014correcting}, which fails on subtle symmetries, so-called doppelgangers, since learned matchers are trained on covisible pairs only.
VidMap~\citep{pataki2026vidmap} resolves doppelgangers by enforcing sequential constraints, which are absent in unordered collections.
Dedicated models~\citep{cai2023doppelgangers, xiangli2025doppelgangers++} predict the covisibility of a pair after matching, but add computational overhead, generalize poorly beyond their training data, and are evaluated on outdoor landmarks only.
In contrast, FlexGlue predicts point-level matches and pair-level covisibility within a single model, benefiting from joint estimation and, following the adaptive depth of LightGlue~\citep{lindenberger2023lightglue}, exits early on unmatchable pairs.

\newcommand{\imga}{\mathbf{I}_a}
\newcommand{\imgb}{\mathbf{I}_b}
\newcommand{\kptsa}{\mathcal{K}_a}
\newcommand{\kptsb}{\mathcal{K}_b}
\newcommand{\corrab}{\mathbf{p}_{a \to b}}
\newcommand{\corrba}{\mathbf{p}_{b \to a}}
\newcommand{\numkptsa}{N}
\newcommand{\numkptsb}{M}
\newcommand{\numlayers}{L}
\newcommand{\numtokens}{P}
\newcommand{\patchsize}{8}
\newcommand{\toksa}{\mathbf{T}_a}
\newcommand{\toksb}{\mathbf{T}_b}
\newcommand{\tokb}{\mathbf{t}_b}
\newcommand{\tokdim}{D}
\newcommand{\querytoken}{\mathbf{q}}
\newcommand{\temperature}{\tau}
\newcommand{\kpt}{\mathbf{k}}
\newcommand{\gridpt}{\mathbf{g}}
\newcommand{\patchwin}{p}
\newcommand{\offset}{\mathbf{\Delta}}
\newcommand{\finedim}{D_f}
\newcommand{\finemap}{\mathbf{F}}
\newcommand{\distab}{\mathbf{D}_{a \to b}}
\newcommand{\distba}{\mathbf{D}_{b \to a}}
\newcommand{\covis}{\bar{o}}
\newcommand{\matchability}{m}
\newcommand{\confid}{c}
\newcommand{\losscoarse}{\mathcal{L}_\text{coarse}}
\newcommand{\lossrobust}{\mathcal{L}_\text{robust}}
\newcommand{\lossfine}{\mathcal{L}_\text{fine}}
\newcommand{\lossmatch}{\mathcal{L}_\text{match}}
\newcommand{\losscovis}{\mathcal{L}_\text{covis}}
\newcommand{\losstotal}{\mathcal{L}}

\begin{figure*}[t]
  \centering
  \begin{minipage}[t]{0.333\linewidth}\centering\small features\end{minipage}%
  \begin{minipage}[t]{0.333\linewidth}\centering\small matches\end{minipage}%
  \begin{minipage}[t]{0.333\linewidth}\centering\small matches after filtering\end{minipage}\\
  \includegraphics[width=\linewidth]{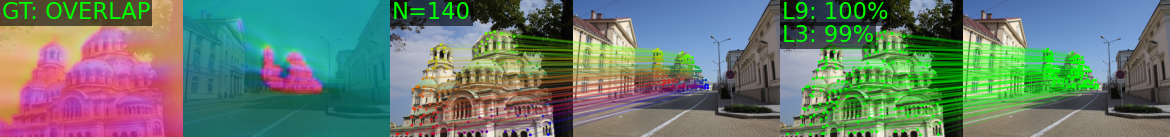}
  \includegraphics[width=\linewidth]{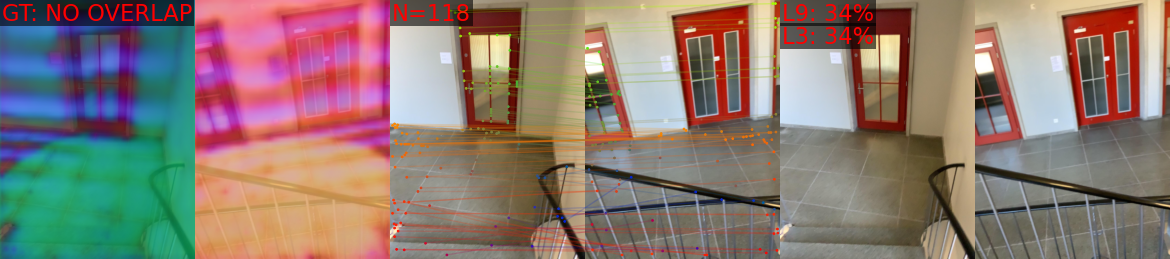}
  \caption{Visual disambiguation examples. \textbf{Left:} Images overlaid with PCA of learned features. \textbf{Center:} Estimated correspondences pre-filtering. \textbf{Right:} We report the covisibility score after 3 and 9 layers in the top-left. \textbf{Top:} Our method correctly identifies a covisible pair. \textbf{Bottom:} We correctly identify challenging doppelgangers from various domains early and stop matching.}
  \label{fig:overlap}
\end{figure*}

\section{Method}

\paragraph{Problem formulation:}
Given two images $\imga, \imgb \in \mathbb{R}^{3 \times H \times W}$ and their corresponding $N$ and $M$ keypoints, with coordinates $\kptsa \in \mathbb{R}^{\numkptsa \times 2}$ and $\kptsb \in \mathbb{R}^{\numkptsb \times 2}$, we aim to estimate two sets of sparse-to-dense correspondences.
Concretely, for each keypoint in $\imga$, FlexGlue predicts its corresponding 2D coordinates $\corrab\in\mathbb{R}^2$ in $\imgb$, and vice versa, $\corrba$.
FlexGlue also predicts whether each point is matchable -- a semantic signal indicating whether the point is visible in the other image.
From the estimated correspondences, we derive point-to-point assignments between keypoints via geometric matching based on their distances in the images.
We further estimate a score whether the two images observe the same scene, in spite of \eg, visual symmetries.

\paragraph{Overview:}
As illustrated in \cref{fig:pipeline}, a frozen pretrained CNN extracts features from each image, which we fuse across layers into a coarse grid of tokens.
A stack of $\numlayers$ identical layers updates these tokens with self- and cross-attention, while a covisibility head decides at each layer whether the two images overlap and halts inference early on non-overlapping pairs.
A lightweight head then reads out, for each query keypoint, a correspondence $\corrab$ with confidence $\confid$ and matchability $\matchability$, which a second head refines to subpixel accuracy.

\subsection{Coarse Matching}

\paragraph{Context aggregation:}
We first extract dense features from each image with a ConvNeXt-B~\citep{convnext}, whose weights are trained with DINOv3~\citep{simeoni2025dinov3} and subsequently remain frozen.
As observed by RoMa~\citep{edstedt2023roma,edstedt2026romav2}, this reduces overfitting to the training data and improves robustness, while making training cheaper.

To combine fine-grained local structure with higher-level semantics, we resample all four backbone stages to stride $\patchsize$ and linearly project them into initial token grids $\toksa^0,\toksb^0 \in \mathbb{R}^{\numtokens{\times}\tokdim}$ at normalized coordinates $\gridpt\in\mathbb{R}^{\numtokens{\times}2}$, with $\numtokens{=}\frac{HW}{\patchsize^2}$.
We then update all tokens using $\numlayers$ Transformer layers $\toksa^l, \toksb^l = \mathcal{T}^l(\toksa^{l-1}, \toksb^{l-1})$ that aggregate context within each image through self-attention and across images through cross-attention.
In self-attention, we encode the relative positions of the token coordinates $\gridpt$ with learned rotary embeddings~\citep{roformer}, following LightGlue~\citep{lindenberger2023lightglue}.
Cross-attention uses no positional encoding.
At stride $\patchsize$, these tokens aggregate distinctive context from the entire image pair, but are too coarse for subpixel-accurate matching. We recover this accuracy through the correspondence readout and local refinement, which we describe next.

\paragraph{Correspondence estimation:}
For a query keypoint $\kpt_a^i \in \kptsa$, we interpolate ($[\cdot]$) the token grid and down-project it into a descriptor $\querytoken_a^i{=}W\toksa^\numlayers[\kpt_a^i]$ with a linear layer $W$, which we share with the other image to reduce the cost of the dense correlation and keep matching symmetric.
Given its cosine similarities $s_{ij}{=}\cos(\querytoken_a^i, \tokb^j)$ to all tokens of the other image, we estimate the correspondence $\corrab^i$ from a global softmax, renormalized within the $3{\times}3$ neighborhood $\mathcal{N}(j^*)$ of the nearest-neighbor cell $j^*{=}\operatorname{argmax}_j s_{ij}$:
\begin{align}
    P_{ij} = \operatorname{softmax}_j\!\left(\frac{s_{ij}}{\temperature}\right),
    \qquad
    \corrab^i = \frac{\sum_{j \in \mathcal{N}(j^*)} P_{ij}\,\gridpt_b^j}{\sum_{j \in \mathcal{N}(j^*)} P_{ij}}
    \enspace,
    \label{eq:readout}
\end{align}
where $\tokb^j$ is the $j$-th token of $W\toksb^\numlayers$ with cell center $\gridpt_b^j$, $\temperature$ is a learned temperature, and $\cos(\mathbf{x},\mathbf{y}){=}\nicefrac{\mathbf{x}^\top\mathbf{y}}{(\norm{\mathbf{x}}_2\norm{\mathbf{y}}_2)}$ is the cosine similarity.
Correspondences $\corrba^j$ for keypoints in $\imgb$ are obtained symmetrically.
Empirically, this windowed readout improves accuracy at depth discontinuities, where a global expectation averages between competing candidates, while avoiding the cell discretization of nearest-neighbor matching at negligible cost.
We supervise $P_{ij}$ with a cross-entropy against the bilinear weights of the ground-truth correspondence over its four surrounding cells (\cref{sec:supervision}), such that the network learns to interpolate between cells from context and reaches sub-cell accuracy.
In practice, we replace the softmax with 1.5-entmax~\citep{peters2019sparse} and the cross-entropy with its sparse counterpart, the entmax loss~\citep{blondel2020learning}, which assigns exactly zero probability to unlikely cells and improves sub-pixel accuracy.
Unlike semi-dense matchers~\citep{sun2021loftr,li2025edm}, we thus avoid quantizing correspondences to cells, and unlike sparse matchers~\citep{sarlin2020superglue,lindenberger2023lightglue}, our targets are derived from the exact corresponding pixel rather than a hard assignment between discrete keypoints.

\paragraph{Matchability:}
We also predict a matchability score per query from its interpolated token $\toksa^\numlayers[\kpt_a^i]$ with a linear classification head.
This score indicates whether the point is visible in the other view and accounts for occlusions and limits of the field-of-view.
We define the confidence of a correspondence as $\confid^i = \matchability^i \sum_{j \in \mathcal{N}(j^*)} P_{ij}$, the matchability times the probability mass within the window.
Compared to sparse matchers, the uniform token grid is not biased by the local feature density and removes the need to learn repeatability, an artifact of local feature detectors.

\subsection{Local Refinement}

Since the coarse readout already places well above 80\% of queries within 2\,px of their ground truth, the remaining error is local.
Instead of a separate fine-scale CNN as in dense matchers~\citep{DKM,edstedt2023roma,edstedt2026romav2}, we follow EDM~\citep{li2025edm} and predict a sub-cell offset $\offset^i$ with a small MLP from the tokens of the current layer, interpolated at $\kpt_a^i$ and $\corrab^i$ and concatenated.
As in EDM, the MLP classifies the offset per axis over bins spanning $\pm1$ cell, and we take the expectation, $\corrab^i \leftarrow \corrab^i + \offset^i$.
This requires only two interpolations and an MLP per query and thus scales linearly with the number of queries.
We stop the gradient through the coarse position $\corrab^i$, which is already supervised by $\losscoarse$, and supervise the refined position with a robust loss (\cref{sec:supervision}).
In practice, it shifts correspondences by only 0.6\,px on average, confirming that it corrects residual sub-pixel errors while leaving the coarse matches intact.

\subsection{Visual Disambiguation}
\paragraph{Rationale:}
Maximizing correspondence accuracy is not sufficient in practice: non-overlapping but visually similar pairs (\cref{fig:overlap}), like opposite facades of a symmetric building, yield spurious correspondences that corrupt the view graph and can collapse downstream 3D reconstruction.
Prior work addresses this as a separate post-processing step, conditioned on pre-extracted correspondences~\citep{cai2023doppelgangers} or on a pretrained two-view geometry model~\citep{xiangli2025doppelgangers++}. We instead integrate disambiguation directly into the matcher, motivated by the observation of LightGlue~\citep{lindenberger2023lightglue} that deep matchers identify covisibility already in early layers, which we confirm for our model (\cref{fig:layers}).

\paragraph{Mechanism:}
A linear layer $W_o$ followed by a sigmoid $\sigma$ maps each token of $\toksa^\numlayers$ and $\toksb^\numlayers$ to a covisibility vote $o_j^a, o_j^b \in [0,1]$, which we average into the pair score $\covis = \tfrac{1}{2\numtokens}\sum_j (o_j^a + o_j^b)$.
Voting per token rather than predicting from a single pooled feature provides dense supervision and requires all tokens to agree on the pair label.
We supervise the score at every layer, such that at inference, we stop after layer $k$ once $\covis < t\,\gamma^{\numlayers-k}$, avoiding computation on unmatchable pairs. Here, $t$ is the pair threshold and the decay $\gamma \in (0,1]$ makes early layers stop only on confident non-covisible pairs.

\subsection{Supervision}
\label{sec:supervision}

During training, we sample one query per $16{\times}16$ cell of a uniform grid, displaced by a random sub-cell offset, rather than using detected keypoints.
Since the network is thus never tied to a specific detector, the query keypoints can be freely exchanged at test time.
The coarse loss is the cross-entropy of $P_{ij}$ against the bilinear weights $w_{ik}$ of the ground-truth correspondence over its four surrounding cells, $\losscoarse = -\sum_k w_{ik} \log P_{ik}$, which we replace by its entmax counterpart, the Fenchel--Young loss (\cref{sec:supp_supervision}), in practice.
Both reach their optimum at $P_{ik}{=}w_{ik}$, for which \cref{eq:readout} recovers the ground-truth position exactly.
$\losscoarse$ and $\lossfine$ are only applied to queries visible in the other image, and thus vanish on non-covisible pairs.
Interpolating in log space is crucial, as minimizing $-\log\sum_k w_{ik} P_{ik}$ instead concentrates all mass on the closest cell and loses sub-cell supervision.
The refinement is supervised with the generalized Charbonnier loss~\citep{barron2019general,edstedt2023roma}.
We only apply it to targets within the $\pm1$-cell range of the offset.
The matchability is supervised per query point with a balanced binary cross-entropy loss $\lossmatch$ for which the label indicates whether the point is visible in the other image.
Similarly, we supervise the per-token covisibility votes with a balanced binary cross-entropy loss $\losscovis$ given the ground-truth pair label.
The total loss at each layer is $\losstotal = 10\,\losscoarse + \lossfine + \lossmatch$, and we add $\losscovis$ once negatives are introduced, with deep supervision across all $\numlayers$ layers. For sparse-to-sparse applications like SfM, we perform 2D nearest neighbor search with predicted correspondences. More details are in the supplementary material.

\subsection{Indoor Doppelgangers Benchmark}

\begin{figure}[t]
  \centering
  \includegraphics[width=0.245\linewidth]{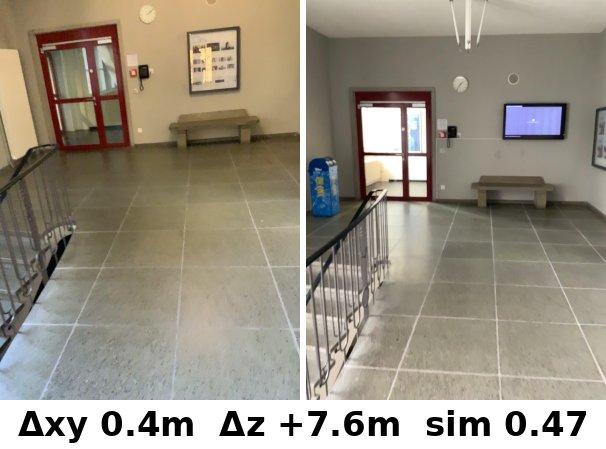}\hfill
  \includegraphics[width=0.245\linewidth]{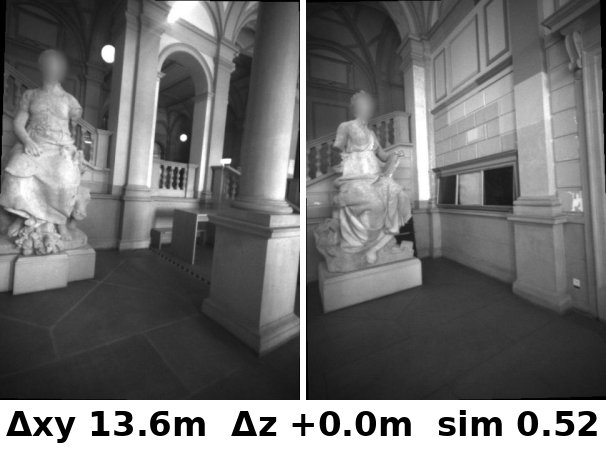}\hfill
  \includegraphics[width=0.245\linewidth]{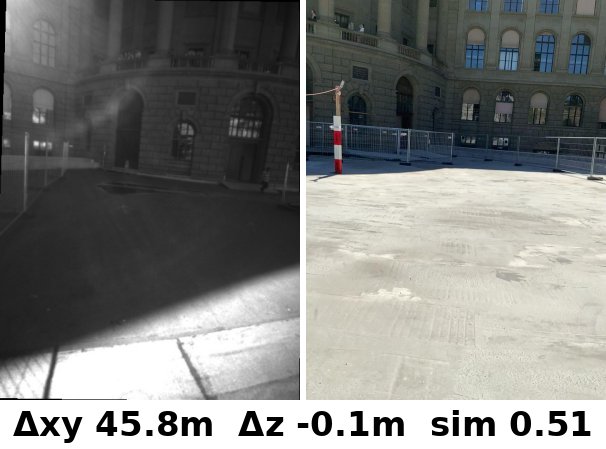}\hfill
  \includegraphics[width=0.245\linewidth]{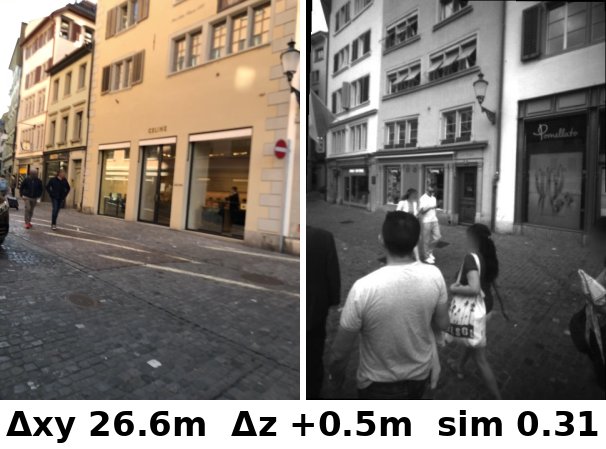}
  \caption{\textbf{Hard negatives of our LaMAR doppelganger benchmark.}
  From left to right: different floors (CAB), indoor (HGE), outdoor (HGE), and cross-device (LIN) pairs.
  None of the pairs overlap, despite a high retrieval similarity (sim) from MegaLoc; $\Delta xy$ and $\Delta z$ denote the horizontal and vertical distance between the two cameras.
  }
  \label{fig:lamar_doppelgangers}
\end{figure}

\paragraph{Motivation:}
Existing doppelganger benchmarks~\citep{cai2023doppelgangers,xiangli2025doppelgangers++} consist exclusively of outdoor scenes.
Indoor environments, however, exhibit their own, particularly hard failure cases: different floors or hallways of the same building often look nearly identical, while repetitive textures, low visual diversity, challenging lighting, and small scene depth further limit the cues available for disambiguation.
To evaluate this so far unexplored setting, we introduce a doppelganger benchmark that includes indoor pairs.

\paragraph{Labels:}
We build the benchmark on three large scenes of LaMAR~\citep{sarlin2022lamar} (CAB, HGE, LIN), which provides accurate ground-truth poses and meshes.
We compute the overlap of each image pair by reprojecting depth rendered from the meshes with the ground-truth poses, and label a pair as positive if its overlap is 10--80\% in both directions and as negative if it is zero (\cref{sec:supp_lamar}). Randomly sampled non-overlapping pairs are trivially rejected. We therefore mine hard negatives, \cref{fig:lamar_doppelgangers}, with the off-the-shelf image retrieval model MegaLoc~\citep{Berton_2025_CVPR}, selecting visually similar yet non-overlapping pairs.
This mirrors practical SfM pipelines, which shortlist candidate pairs by retrieval before matching.
As positives, we include both high-similarity pairs, such that the retrieval score alone does not separate the classes, and pairs with very low overlap, which are critical to retain for loop closure.
Each scene contains 1000 pairs, 500 positive and 500 negative.

\paragraph{Availability of training data:}
Collecting indoor hard negatives at scale is difficult: unlike internet photo collections, indoor images carry no GPS tags that would reveal visually similar but distinct places.
Consequently, all existing training sets for visual disambiguation are outdoor, and we evaluate all methods zero-shot to measure how well they generalize across this domain gap.
As shown in \cref{tab:doppelgangers}, all methods drop substantially compared to the pure outdoor benchmarks. In \cref{tab:ablation_dg}, we build a small training dataset from NaverLabs~\citep{lee2021naverlabs} and show that this helps in-domain generalization.

\section{Experiments}
\label{sec:experiments}

\subsection{Implementation Details}
We train FlexGlue on eight datasets: MegaDepth~\citep{li2018megadepth}, BlendedMVS~\citep{yao2020blendedmvs}, ScanNet++~\citep{yeshwanth2023scannet++}, the TartanAir~\citep{tartanair2020iros} split of MapAnything~\citep{keetha2026mapanything}, Hypersim~\citep{hypersim}, Map-free~\citep{mapfree}, AerialMegaDepth~\citep{aerialmegadepth} and DL3DV~\citep{dl3dv} with balanced sampling over scenes.
For all datasets we sample pairs with dense visual overlap, normalized by area with depth coverage, in $[0.05, 0.8]$.
Images are resized to $512{\times}512$\,px.
We train for 50 epochs with 150k samples per epoch, a batch size of 16 per device, and mixed precision on 4 H200 GPUs ($\sim$4 GPU-days).
With gradient checkpointing the same model can be trained on 4 RTX 4090 in about 1.5 days.
We use a learning rate of $2\cdot10^{-4}$, held constant for 5 epochs, then decayed to $2 \cdot 10^{-5}$ over the remaining 45 epochs.
The matching temperature is initialized to 10 and learned independently per layer.
We use the ConvNeXt-B backbone of DINOv3, a transformer with 9 layers and an internal dimension of 512, a 64-dimensional matching projection $W$, and 16 offset bins per axis in the refinement head.
We apply light photometric augmentations, and additionally continuous rotation and perspective augmentations in the 10-epoch fine-tuning stage.
In the latter, we reset the learning rate to $5\cdot10^{-5}$ and mix in Doppelgangers~\citep{cai2023doppelgangers} pairs (50\% positive, 50\% negative) and flip augmentations to generate additional hard negatives.
At inference, we use a pair threshold $t{=}0.9$ and decay $\gamma{=}0.9$.

\subsection{Relative Pose Estimation}

\paragraph{Setup:}
We evaluate image matching via calibrated relative pose estimation on two established benchmarks: MegaDepth-1500~\citep{sun2021loftr}, consisting of outdoor phototourism image pairs, and ScanNet-1500~\citep{dai2017scannet,sarlin2020superglue}, an indoor benchmark with challenging self-similarities and textureless regions.
We further evaluate all methods on the challenging HardMatch benchmark~\citep{nordstrom2026loma} which contains image pairs under extreme appearance variations. We provide additional details in the supplementary materials.

\paragraph{Results:}
As reported in \cref{tab:main-results},
in the sparse setting, FlexGlue performs similar to LoMa~\citep{nordstrom2026loma}.
The large improvement on indoor scenes -- where sparse matchers traditionally struggle due to repetitive textures and limited visual diversity -- demonstrates that grid-based context aggregation is more effective there than between sparse keypoints.
In the sparse-to-dense setting, our method performs competitively with RoMav2 while being substantially faster.
Notably, dense correspondence estimation consistently improves over the sparse-to-sparse variant on relative pose estimation, but lacks robustness on HardMatch.
Sparse-to-Dense methods take the middle-ground, and our method achieves highest accuracy on HardMatch.

\begin{table}[h!]
  \caption{\textbf{Relative pose estimation on MegaDepth-1500, ScanNet-1500 and HardMatch.}
  We report the AUC of the pose error at $5^\circ$, $10^\circ$, and $20^\circ$, the HardMatch mAA@10px, and the time for extraction and matching.
  We highlight the \colorbox{rankfirst}{first}, \colorbox{ranksecond}{second}, and \colorbox{rankthird}{third} best results.
  FlexGlue has comparable accuracy with dense matchers at a fraction of their cost, and also improves sparse-to-sparse feature matching on ScanNet and HardMatch.}
  \label{tab:main-results}
  \centering
  \footnotesize
  \setlength{\tabcolsep}{2.5pt}
  \begin{tabular}{@{}llllccccccccc@{}}
    \toprule
    & & \multirow{2}{*}{Detector} & \multirow{2}{*}{Matching} & \multicolumn{3}{c}{MegaDepth-1500} & \multicolumn{3}{c}{ScanNet-1500} & HardMatch & Extr.& Match \\
    \cmidrule(lr){5-7} \cmidrule(lr){8-10} \cmidrule(lr){11-11} \cmidrule(lr){12-12} \cmidrule(lr){13-13}
    & & & & $5^\circ$ & $10^\circ$ & $20^\circ$ & $5^\circ$ & $10^\circ$ & $20^\circ$ & mAA@10px & [ms] & [ms] \\
    \midrule
    \cellcolor{groupgray} & \cellcolor{groupgray} & ALIKED & LightGlue & \trd 52.4 & \trd 68.8 & \trd 81.0 & \trd 17.0 & \trd 34.2 & \trd 51.3 & \trd 36.3 & \snd 36.0 & \fst 21.7 \\
    \cellcolor{groupgray} & \cellcolor{groupgray} & DaD & LoMa-G & \fst 56.1 & \fst 72.2 & \fst 84.0 & \snd 29.3 & \snd 51.7 & \snd 70.0 & \snd 54.3 & \trd 128.5 & \trd 49.8 \\
    \multirow{-3}{*}{\cellcolor{groupgray}\rotatebox[origin=c]{90}{\textit{Sparse}}} & \multirow{-3}{*}{\cellcolor{groupgray}\rotatebox[origin=c]{90}{\scriptsize 4096 kp.}} & SuperPoint & \textbf{Ours} & \snd 53.8 & \snd 70.0 & \snd 82.0 & \fst 31.9 & \fst 54.8 & \fst 72.6 & \fst 55.7 & \fst 33.1 & \snd 29.6 \\
    \midrule
    \cellcolor{groupgray} & \cellcolor{groupgray} & - & RoMa & \snd 62.6 & \snd 76.7 & \snd 86.3 & \trd 31.8 & 53.4 & 70.9 & \snd 48.1 & 49.4 & 239.5\\
    \cellcolor{groupgray} & \cellcolor{groupgray} & - & MASt3R & 54.5 & 71.9 & 84.0 & \fst 34.0 & \fst 57.0 & \fst 74.4 & 33.5 & 51.1 & 747.2 \\
    \cellcolor{groupgray} & \cellcolor{groupgray} & - & EDM & 57.5 & 73.2 & 84.2 & 19.8 & 37.5 & 54.4 & 33.1 & \fst 4.0 & \fst 15.4 \\
    \cellcolor{groupgray} & \cellcolor{groupgray} & - & UFM & 41.5 & 57.9 & 72.4 & 31.3 & 54.1 & \trd 72.0 & 33.9 & \trd 35.7 & \trd 37.4 \\
    \cellcolor{groupgray} & \cellcolor{groupgray} & - & RoMav2 & \fst 62.8 & \fst 77.0 & \fst 86.6 & \snd 33.6 & \trd 56.2 & \snd 73.8 & \trd 46.5 & 42.4 & 73.1 \\
    \multirow{-6}{*}{\cellcolor{groupgray}\rotatebox[origin=c]{90}{\textit{Dense}}} & \multirow{-6}{*}{\cellcolor{groupgray}--} & & \textbf{Ours} & \trd 58.1 & \trd 73.4 & \trd 84.4 & \snd 33.6 & \snd 56.4 & \snd 73.8 & \fst 55.6 & \snd 11.8 & \snd 28.4 \\
    \midrule
    \cellcolor{groupgray} & \cellcolor{groupgray} & SuperPoint & CAPS & 34.8 & 52.2 & 67.5 & 9.4 & 21.9 & 37.1 & - & - & - \\
    \cellcolor{groupgray} & \cellcolor{groupgray} & SuperPoint & S2DNet & 35.0 & 51.9 & 64.2 & 10.3 & 21.6 & 34.5 &  - & - & - \\
    \cellcolor{groupgray} & \cellcolor{groupgray} & SuperPoint & RCM & 53.2 & 69.4 & 81.5 & \trd 17.3 & 34.6 & 52.1 & - & - & - \\
    \cellcolor{groupgray} & \cellcolor{groupgray} & SuperPoint & RoMav2 & \snd 57.1 & \snd 73.1 & \snd 84.1 & \snd 33.5 & \trd 56.0 & \trd 73.3 & \trd 49.0 & \snd 53.4 & \trd 63.2 \\
    \cellcolor{groupgray} & \cellcolor{groupgray} & SuperPoint & \textbf{Ours} & \trd 55.8 & \trd 72.0 & \trd 83.6 & \snd 33.5 & \fst 56.5 & \fst 73.9 & \snd 56.4 & \fst 31.0 & \fst 28.7 \\
    \multirow{-6}{*}{\cellcolor{groupgray}\rotatebox[origin=c]{90}{\textit{Sparse-to-Dense}}} & \multirow{-6}{*}{\cellcolor{groupgray}\rotatebox[origin=c]{90}{\scriptsize 2048 keypoints}} & & \quad+RoMav2 refine & \fst 59.4 & \fst 75.0 & \fst 85.7 & \fst 33.7 & \snd 56.2 & \snd 73.8 & \fst 57.2 & \fst 31.0 & \snd 46.8 \\
    \bottomrule
  \end{tabular}
\end{table}

\subsection{Two-View Estimation}

\paragraph{Setup:}
We perform a diverse evaluation on seven two-view estimation benchmarks.
Ground-level pairs come from HPatches~\citep{balntas2017hpatches} (homography AUC), WxBS~\citep{mishkin2015wxbs} (epipolar mAA under extreme appearance and modality changes), RUBIK~\citep{loiseau2025rubik} (pose AUC and success rate on driving scenes), and IMC22~\citep{imwchallenge2022} (pose mAA on phototourism).
TerraSky3D~\citep{durso2026terrasky3d} pairs ground-level with drone images (100 pairs from each of 29 scenes, pose AUC), while EarthMatch~\citep{berton2024earthmatch} and SatAst with rotations ~\citep{edstedt2026romav2} match astronaut photographs to satellite images (localization rate and homography transfer AUC of annotated points). We use the default settings for all baselines.

\paragraph{Results:}
As reported in \cref{tab:additional}, FlexGlue with 2048 DaD~\citep{edstedt2025dad} keypoints in sparse-to-dense mode achieves the highest accuracy on all benchmarks but IMC22, outperforming the sparse matchers ALIKED+LightGlue~\citep{Zhao2023ALIKED,lindenberger2023lightglue} and LoMa-G~\citep{nordstrom2026loma} as well as the dense matcher RoMav2~\citep{edstedt2026romav2}.
The gains are largest on aerial imagery, where FlexGlue even outperforms the rotation-invariant LoMa-R~\citep{nordstrom2026orientation}.
Increasing the input resolution yields consistent gains across benchmarks, and an additional refinement with RoMav2~\citep{edstedt2026romav2} boosts the accuracy to 67.3 AUC@10\,px on SatAst and 66.6 AUC@$10^\circ$ on TerraSky3D.

\begin{table}[t]
  \caption{\textbf{Two-view matching in additional domains.}
  FlexGlue exhibits the highest matching accuracy over multiple datasets from different domains.
  A higher resolution further improves results.
  }
  \label{tab:additional}
  \centering
  \footnotesize
  \setlength{\tabcolsep}{2.3pt}
  \begin{tabular}{@{}lcccccccccccc@{}}
    \toprule
    & \multicolumn{8}{c}{Ground--Ground} & \multicolumn{2}{c}{Ground--Aerial} & \multicolumn{2}{c}{Aerial--Aerial} \\
    \cmidrule(lr){2-9} \cmidrule(lr){10-11} \cmidrule(lr){12-13}
    \multirow{2}{*}{Method} & \multicolumn{3}{c}{HPatches} & WxBS & \multicolumn{3}{c}{RUBIK} & IMC22 & \multicolumn{2}{c}{TerraSky3D} & EarthMatch & SatAst \\
    \cmidrule(lr){2-4} \cmidrule(lr){5-5} \cmidrule(lr){6-8} \cmidrule(lr){9-9} \cmidrule(lr){10-11} \cmidrule(lr){12-12} \cmidrule(lr){13-13}
    & 3px & 5px & 10px & mAA\kern-0.12em@\kern-0.12em10px & $10^\circ$ & $20^\circ$ & succ. & mAA & $10^\circ$ & $20^\circ$ & \% loc. & 10px \\
    \midrule
    ALIKED+LG & 66.2 & 76.9 & 86.3 & 43.9 & 52.8 & 59.1 & 39.0 & 79.2 & 25.3 & 29.4 & 84.0 & 18.4 \\
    LoMa-G & 66.4 & 77.5 & 87.3 & \snd 73.4 & 73.1 & 79.8 & \trd 47.6 & \fst 89.8 & 53.4 & 61.6 & \trd 98.0 & 24.4 \\
    LoMa-R & \trd 66.8 & 77.9 & 87.7 & 65.5 & 68.6 & 75.9 & 45.1 & 86.3 & 44.2 & 51.7 & \fst 99.6 & \trd 48.3 \\
    RoMav2 & \snd 72.2 & \trd 81.2 & \trd 89.2 & 64.8 & \trd 78.4 & \trd 86.2 & \fst 59.5 & 84.7 & \trd 53.9 & \trd 62.9 & 92.2 & 38.1 \\
    \textbf{Ours} & \fst 72.9 & \snd 81.9 & \snd 89.8 & \trd 72.6 & \snd 81.9 & \snd 88.3 & \snd 58.7 & \trd 87.0 & \snd 56.9 & \snd 67.7 & \snd 99.2 & \snd 48.8 \\
    \textbf{Ours} (640 px) & \fst 72.9 & \fst 82.0 & \fst 89.9 & \fst 73.7 & \fst 83.2 & \fst 89.2 & \fst 59.5 & \snd 88.5 & \fst 62.0 & \fst 72.4 & \fst 99.6 & \fst 60.8 \\
    \bottomrule
  \end{tabular}
\end{table}

\subsection{Visual Disambiguation}

\paragraph{Setup:}
We consider three benchmarks: Doppelgangers~\citep{cai2023doppelgangers}, which contains phototourism image pairs that often require analyzing fine-grained structural details, ViSym~\citep{xiangli2025doppelgangers++}, a collection of phone-captured street scene pairs, for which we use our cleaned test set ViSym-F (\cref{sec:supp_visym}), and our LaMAR-DG benchmark.
We report AUROC, average precision (AP) and run time per method, and compare against Doppelgangers~\citep{cai2023doppelgangers} and DG++~\citep{xiangli2025doppelgangers++}. Further details in appendix.

\paragraph{Results:}
As reported in \cref{tab:doppelgangers},
FlexGlue achieves the highest accuracy on ViSym-F, outperforming DG++ while being over $4\times$ faster.
On the in-distribution Doppelgangers benchmark, FlexGlue performs on par with DG++.
On out-of-distribution data, FlexGlue outperforms SotA methods, especially on the extremely challenging LaMAR benchmark.
Notably, our method outperforms literature even without any manually annotated doppelganger training pairs, suggesting that matching data already provides most of the supervision needed for disambiguation (\cref{tab:ablation_dg}).

\begin{table}[!t]
  \caption{%
  \textbf{Detection of Doppelgangers.}
  We report AUROC and AP on DG (in-distribution), ViSym and LaMAR, as well as the matching time.
  FlexGlue achieves the highest disambiguation accuracy across all datasets when trained on the same labelled data (DG), especially +10 AUROC points on the LaMAR benchmark, at much higher throughput.
  }
  \label{tab:doppelgangers}
  \centering
  \footnotesize
  \setlength{\tabcolsep}{4.5pt}  %
  \begin{tabular}{@{}llccccccc@{}}
    \toprule
    \multirow{2}{*}{Method} & \multirow{2}{*}{Labelled data} & \multicolumn{2}{c}{DG} & \multicolumn{2}{c}{ViSym-F} & \multicolumn{2}{c}{LaMAR} & Time \\
    \cmidrule(lr){3-4} \cmidrule(lr){5-6} \cmidrule(lr){7-8}
    & & AUROC & AP & AUROC & AP & AUROC & AP & [ms] \\
    \midrule
    DG~\citep{cai2023doppelgangers} & DG & 94.4 & 95.4 & 66.2 & 74.1 & 69.5 & 69.8 & \trd 139.5 \\
    DG++~\citep{xiangli2025doppelgangers++} & DG & \snd 98.1 & \trd 98.0 & 86.4 & \trd 90.9 & \trd 78.6 & \trd 79.4 & \snd 124.7 \\
    DG++~\citep{xiangli2025doppelgangers++} & DG + ViSym & \snd 98.1 & \snd 98.1 & \trd 88.3 & 90.7 & 73.3 & 75.3 & \snd 124.7 \\
    Ours & matching & \trd 97.0 & 97.1 & \snd 95.3 & \snd 96.5 & \snd 88.5 & \snd 89.0 & \fst 28.7 \\
    Ours & DG + matching & \fst 98.4 & \fst 98.2 & \fst 95.9 & \fst 96.8 & \fst 88.8 & \fst 89.2 & \fst 28.7 \\  %
    \bottomrule
  \end{tabular}
\end{table}

\subsection{Multi-view estimation}

\paragraph{Setup:}
We evaluate end-to-end SfM on 25 scenes of the ETH3D dataset~\citep{schops2017multi}, spanning both indoor and outdoor environments.
The main challenges of this dataset are low-texture environments, self-similarities, and symmetries, which are classical failure cases of SfM.
This benchmark is a joint test of both matching accuracy and visual disambiguation.
We report the AUC of relative pose errors within each scene, averaged over all scenes, as well as the matching throughput as measured by hloc~\citep{sarlin2019coarse}.
We also evaluate noisy 3D reconstructions on the CVPR 2025 image matching challenge~\citep{image-matching-challenge-2025}.
This benchmark provides random collections of images, and requires grouping and accurately reconstructing multiple clusters in one scene.
We run all methods on the hidden test set on Kaggle~\citep{image-matching-challenge-2025-ongoing, image-matching-challenge-2025}, and report the private and public score, and the score on the ongoing benchmark there.

\paragraph{Results:}
As shown in \cref{tab:eth3d}, even without visual disambiguation, our method (SP+Ours) outperforms all other methods without filtering.
All deep feature matchers break under symmetries, including dense RoMa v2.
Pairing image matchers with visual disambiguation significantly increases the robustness of all methods, but at lower throughput.
Notably, early stopping hardly impairs accuracy while increasing throughput further, avoiding wasteful computation on unmatchable pairs, and still outperforms the combined RoMav2+DG++ baseline by a comfortable margin at a fraction of its runtime.
These results support the central claim that robust multi-view reconstruction requires both high matching accuracy and effective visual disambiguation, which our model provides jointly.

On IMC2025 (\cref{tab:imc2025}), our method outperforms the heavily tuned 2025 winning MASt3R~\citep{leroy2024grounding} entry with 4096 DaD~\citep{edstedt2025dad} keypoints only.
With visual disambiguation we improve 4 points over matching, and 9 points over LoMa~\citep{nordstrom2026loma} with the same keypoints.
On the ongoing benchmark~\citep{image-matching-challenge-2025-ongoing}, FlexGlue with 10k keypoints outperforms the top leaderboard entry by 8 points without extensive tuning, while LoMa exceeds the Kaggle time limit because of slower matching at this budget (see~\cref{fig:runtime-a}).

\begin{table}[!t]
  \setlength{\tabcolsep}{4pt}
  \begin{minipage}[t]{0.53\linewidth}
      \caption{\textbf{Uncalibrated end-to-end SfM.}
 On ETH3D, FlexGlue exhibits a higher matching accuracy than existing approaches, resulting in more accurate poses.
  Disambiguation further increases pose accuracy by preventing collapse of reconstructions. %
  }
  \label{tab:eth3d}
  \centering
  \scriptsize
  \setlength{\tabcolsep}{2.5pt}
  \begin{tabular}{@{}lllcccc@{}}
    \toprule
    \multirow{2}{*}{Keypoints} & \multirow{2}{*}{Matching} & \multirow{2}{*}{Filter} & \multicolumn{3}{c}{Pose AUC} & Throughput \\
    \cmidrule(lr){4-6}
      & & & $1^\circ$ & $3^\circ$ & $5^\circ$ & [pairs/s] \\
    \midrule
    SuperPoint & LightGlue & - & 29.5 & 51.8 & 58.5 & \fst 40 \\
    SuperPoint & RoMav2 & - & \trd 56.4 & \trd 75.7 & \trd 80.7 & 11 \\
     DaD & LoMa-B & - & \snd 58.3 & \snd 77.6 & \snd 82.3 & \snd 37 \\
     SuperPoint & Ours & - & \fst 60.1 & \fst 78.9 & \fst 83.5 & \trd 30 \\
    \midrule
    - & MASt3R-SfM & DG++ & 54.1 & 77.9 & 81.5 & \trd 5 \\
    SuperPoint & RoMav2 & DG++ & 62.0 & 80.6 & 85.1 & 4 \\
    DaD & LoMa-B & DG++ & \trd 62.4 & \trd 81.2 & \trd 85.8 & 4 \\
    SuperPoint & Ours & Ours & \fst 68.9 & \fst 86.8 & \fst 92.3 & \snd 30 \\
    SuperPoint & Ours & Ours (early) & \snd 67.5 & \snd 86.0 & \snd 91.6 & \fst 45 \\
    \bottomrule
  \end{tabular}

  \end{minipage}\hfill
  \begin{minipage}[t]{0.44\linewidth}
      \caption{\textbf{SfM with distractors.} On IMC2025, FlexGlue yields better reconstructions thanks to its disambiguation.
  Using 10k keypoints instead of 4k further improves, but makes LoMa timeout on Kaggle (OOT).
  }
  \label{tab:imc2025}
  \centering
  \scriptsize
  \begin{tabular}{@{}llccc@{}}
    \toprule
    \multirow{2}{*}{Matching} & \multirow{2}{*}{Filter} & Private & Public & Ongoing \\
      & & Score & Score & Score \\
    \midrule
    Top leaderboard & n/a & 56.00 & 52.64 & \trd 65.56 \\  %
    \midrule
    LoMa-B & - & 57.01 & 53.34 & 55.67 \\
    LoMa-G & - & 49.38 & 46.17 & 46.76 \\
    LoMa-R & - & 44.15 & 40.83 & 42.38 \\
    Ours & - & \trd 61.74 & \trd 60.86 & 61.23 \\
    Ours & Ours & \snd 66.04 & \snd 65.18 & \snd 65.65 \\
    \midrule
    LoMa-B (10k) & - & OOT & OOT & OOT \\
    Ours (10k) & Ours & \fst 74.62 & \fst 73.78 & \fst 74.07 \\  %
    \bottomrule
  \end{tabular}

  \end{minipage}
\end{table}

\subsection{Visual Localization}

We evaluate indoor visual localization on InLoc~\citep{taira2018inloc}, where the goal is to estimate the 6-DoF absolute pose of a query image given a set of database images with known poses and dense depth maps.
We follow the hloc~\citep{sarlin2019coarse} protocol.
Further setup details can be found in supplementary.
Dense and semi-dense matchers are less affected by keypoint repeatability and generally outperform sparse matching on InLoc (\cref{tab:inloc}), although the recent LoMa-G narrows the gap considerably.
Our method is more robust and as precise as dense matchers, achieving the best results on 5 of 6 thresholds, at much higher throughput.

\begin{table}[t]
  \begin{minipage}[t]{0.53\linewidth}
    \vspace{0pt}
      \setlength{\tabcolsep}{2.5pt}
  \caption{%
  \textbf{Visual localization on InLoc.}
  We report the ratio of correctly localized queries at three error thresholds and the matching throughput. FlexGlue is most accurate and has competitive efficiency.}
  \label{tab:inloc}
  \centering
  \footnotesize
  \begin{tabular}{@{}lccccccc@{}}
    \toprule
    \multirow{2}{*}{Method} & \multicolumn{3}{c}{DUC1} & \multicolumn{3}{c}{DUC2} & matching \\
    \cmidrule(lr){2-4} \cmidrule(lr){5-7}
    & 0.25m & 0.5m & 1m & 0.25m & 0.5m & 1m & [pairs/s] \\
    \midrule
    SP+SG & 47.0 & 69.2 & 79.8 & 53.4 & 77.1 & 80.9 & 25 \\
    SP+LG & 49.0 & 68.2 & 79.3 & 55.0 & 74.8 & 79.4 & \snd  35 \\

    RoMa & \trd 60.6 & \trd 79.3 & 89.9 & 66.4 & 83.2 & 87.8 & 4 \\
    MASt3R & 56.1 & \trd 79.3 & \trd 90.9 & \trd 71.0 & \trd 87.0 & \snd 91.6 & - \\
    EDM & 51.5 & 72.7 & 85.9 & 59.5 & 82.4 & 88.5 & \fst  40 \\
    RoMav2 & \fst 67.7 & \fst 84.3 & \fst 94.4 & \snd 73.3 & 86.3 & \trd 90.8 &  8 \\
    LoMa-G & 55.6 & \snd 80.3 & \snd 91.4 & \snd 73.3 & \snd 87.8 & 89.3 & 18 \\
    Ours & \snd 62.6 & \fst 84.3 & \fst 94.4 & \fst 77.9 & \fst 88.5 & \fst 94.7 & \trd 32 \\
    \bottomrule
  \end{tabular}

  \end{minipage}\hfill
  \begin{minipage}[t]{0.45\linewidth}
    \vspace{0pt}
      \caption{\textbf{Matching ablation.} HardMatch measures robustness, PCK@1\,px precision. The refinement head improves sub-pixel precision, and 1.5-entmax increases robustness.}
  \label{tab:ablation}
  \centering
  \scriptsize
  \setlength{\tabcolsep}{2pt}
  \begin{tabular}{@{}lccc@{}}
    \toprule
    \multirow{2}{*}{Component} & HardMatch & EPE & Match \\
     & mAA@10px & PCK@1px & [ms] \\
    \midrule
    Ours & \snd 56.4 & 61.0 & 28.7 \\
    \midrule
    \ 3 layers & 49.9 & 39.3 & \fst 16.6 \\
    \ w/o refinement head & \trd 55.5 & 53.8 & \trd 27.0 \\
    \ w/o windowed soft-argmax & 52.8 & 59.1 & \snd 26.8 \\
    \ w/o disambiguation finetune & 54.8 & \trd 62.0 & 28.7 \\
    \ softmax instead of entmax & 48.6 & 55.5 & 27.6 \\
    \midrule
    w/ RoMav2 refine & \fst 57.2 & \snd 64.0 & 46.8 \\
    \midrule
    RoMav2 & 49.0 & \fst 65.6 & 63.2 \\  %
    RoMav2 w/o refinement & 41.6 & 26.3 & 39.1 \\
    \bottomrule
  \end{tabular}

  \end{minipage}
\end{table}
\begin{table}[t]
  \centering
  \caption{\textbf{Doppelgangers ablation.} AUROC on ViSym-F and LaMAR Disambiguation. Our model learns disambiguation early, and guidance from match supervision improves generalization. Even without labelled data our model achieves high accuracy. Training in-domain increases accuracy. Right: Qualitative examples with GT, DG++~\citep{xiangli2025doppelgangers++} and our disambiguation score.}
  \label{tab:ablation_dg}
  \newsavebox{\dgtabbox}\newlength{\dgimgh}%
  \sbox{\dgtabbox}{  \centering
  \scriptsize
  \setlength{\tabcolsep}{2pt}%
  \begin{tabular}{@{}lcc@{}}
    \toprule
    \multirow{2}{*}{Method} & ViSym-F & LaMAR \\
     & AUROC & AUROC \\
    \midrule
    Ours & \snd 95.9 & \snd 88.8\\
    \midrule
    - 3 layers & 93.4 & 87.4 \\
    - no matching loss & 93.4 & 79.5 \\
    - DG data & \trd 95.3 & \trd 88.5 \\
    \midrule
    + NaverLabs & \fst 97.4 & \fst 92.3 \\
    \midrule
    DG++ w/ ViSym & 88.3 & 73.3 \\

    \bottomrule
  \end{tabular}%
}%
  \setlength{\dgimgh}{\dimexpr(\linewidth-\wd\dgtabbox-9pt)*1000/3843\relax}%
  \noindent\parbox[c]{\wd\dgtabbox}{\usebox{\dgtabbox}}\hspace{4pt}%
  \parbox[c]{\dimexpr\linewidth-\wd\dgtabbox-4pt\relax}{%
    \includegraphics[height=\dgimgh]{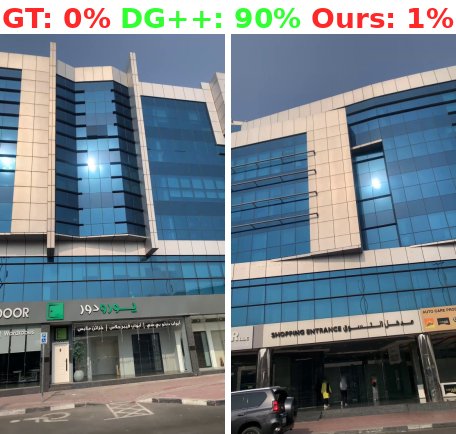}\hspace{2pt}%
    \includegraphics[height=\dgimgh]{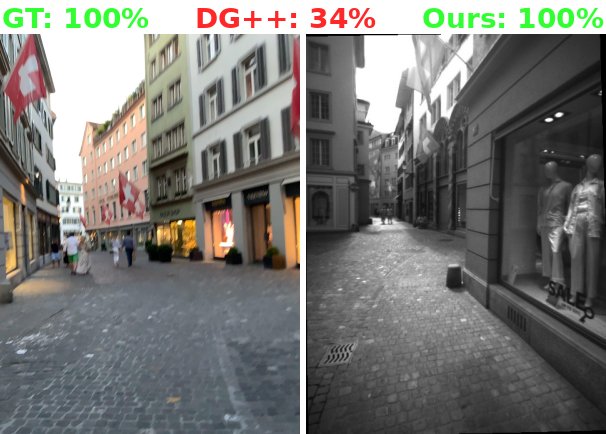}\hspace{2pt}%
    \includegraphics[height=\dgimgh]{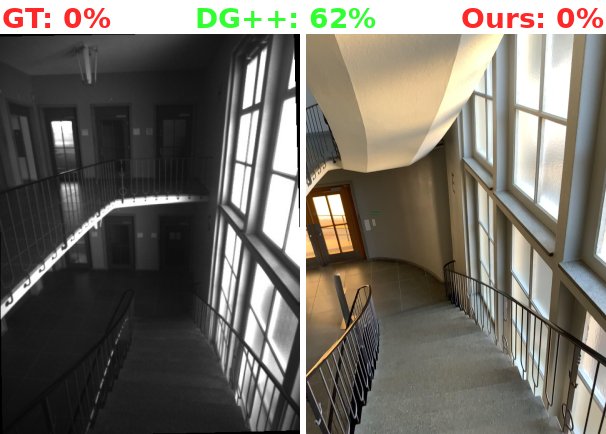}}
\end{table}

\subsection{Ablations}
\paragraph{Matching:}
\Cref{tab:ablation} removes or replaces one component at a time, with SuperPoint~\citep{superpoint} queries.
Replacing 1.5-entmax by a softmax causes the largest drop in robustness and a clear drop in precision.
The windowed soft-argmax mainly improves robustness, while the refinement head improves sub-pixel precision at negligible cost.
Fine-tuning for disambiguation slightly improves robustness at a small cost in precision.
Stopping after 3 layers nearly halves the matching time but costs accuracy (\cref{fig:runtime}).
Unlike RoMav2, whose coarse matches are imprecise without its dense refinement, our coarse readout is already accurate, and adding the RoMav2 refiner further improves precision.

\paragraph{Visual disambiguation:}
\Cref{tab:ablation_dg} ablates the covisibility head on ViSym-F and LaMAR-DG.
Training for covisibility alone, without the matching losses, drops AUROC by up to 9.3 points, confirming that matching and disambiguation reinforce each other.
Without the DG training data, relying only on flip augmentations and negatives from our matching datasets, our model retains most of its performance.
Adding additional indoor negatives mined on NaverLabs~\citep{lee2021naverlabs} further improves LaMAR.

\section{Conclusion}

We introduced a deep matching architecture to obtain sparse-to-dense correspondences which bridges the efficiency of sparse matchers and robustness of dense matchers. By combining uniform attention over a grid with well localized query points and visual disambiguation, our method achieves the highest accuracy on two multi-view reconstruction benchmarks (ETH3D, IMC2025). We further improve over literature on all disambiguation benchmarks, especially the proposed indoor LaMAR dataset, on InLoc visual localization, and the challenging two-view HardMatch benchmark, all while maintaining competitive computational efficiency.

\section{Acknowledgements}

We sincerely thank Mihai Dusmanu for his immense help, feedback and
discussions in the initial phase of the project. This work was supported by the Swiss AI Initiative through a grant from the Swiss National Supercomputing Centre (CSCS) under project ID ab036 on the Alps infrastructure; the Swiss National Science Foundation (Advanced Grant 216260: "Beyond Frozen Worlds: Capturing Functional 3D Digital Twins from the Real World"); the European Union’s Horizon Europe research and innovation programme under grant agreement No. 101214398 (ELLIOT); the Horizon Europe programme under grant agreement No. 101298421 (GRAIL); and the ETH AI Center through an ETH AI Center postdoc fellowship awarded to Sunghwan Hong.

\appendix
\section{Appendix}

We report additional results on inference time, keypoint detectors and LaMAR-DG, describe the construction of our benchmarks, provide implementation and evaluation details, and show qualitative examples.

\section{Additional results}

\subsection{Inference time \& layer ablation}
\begin{figure}[b]
      \centering
      \includegraphics[width=0.49\linewidth]{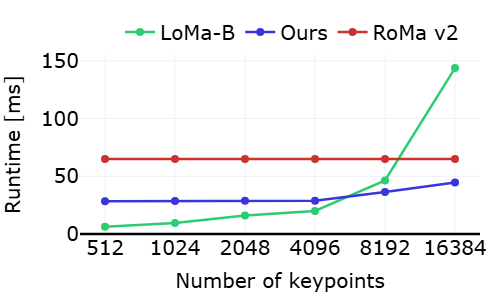}\hfill
      \includegraphics[width=0.49\linewidth]{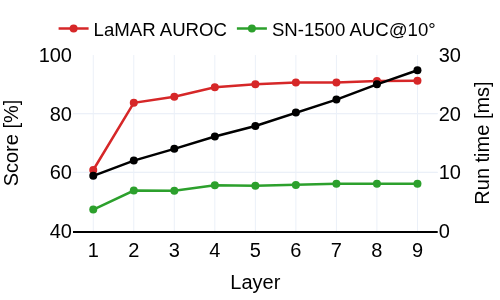}
      \caption{\textbf{Run-time on different keypoint budgets and layer ablations}. RoMav2's~\citep{edstedt2026romav2} run time is independent of the number of keypoints, but has a large overhead. LoMa~\citep{nordstrom2026loma} is very efficient but scales quadratically, while our method is fairly efficient and scales linearly (left). Our method learns strong matching and disambiguation early on, while the run time grows linearly with the number of layers (right, black line).}
      \label{fig:runtime}
      \label{fig:runtime-a}
      \label{fig:layers}
  \end{figure}

\cref{fig:runtime} compares the matching time as a function of the number of keypoints.
All methods use mixed precision, and we measure run time on a single RTX 4090.
Dense matchers such as RoMav2~\citep{edstedt2026romav2} have a constant but large cost.
Sparse matchers such as LightGlue~\citep{lindenberger2023lightglue} and LoMa~\citep{nordstrom2026loma} are fast for few keypoints, but they scale quadratically with keypoints.
FlexGlue aggregates context over a fixed token grid, and only its lightweight readout scales, linearly, with the number of queries, making it faster than sparse matchers for large keypoint sets and faster than RoMav2. In~\cref{fig:runtime}, on the right, we report matching performance and visual disambiguation performance against the layer count. We observe that the model yields strong performance on both benchmarks very early in the network, supporting the early-exit on negative pairs.
Our model takes around 1.2 GB VRAM for extraction and matching (0.7 GB for matching only), against 2.4 GB (0.5 GB) for LoMa-B and 3.8 GB (1.5 GB) for LoMa-G~\citep{nordstrom2026loma}.

\begin{table}[b]
  \caption{\textbf{Keypoint detectors.} FlexGlue works with any detector. RaCo yields the most accurate reconstructions, while all detectors perform similarly on two-view matching.
  }
  \label{tab:detectors}
  \centering
  \footnotesize
  \setlength{\tabcolsep}{4pt}
  \begin{tabular}{@{}lcccccc@{}}
    \toprule
    \multirow{2}{*}{Keypoints} & \multicolumn{4}{c}{ETH3D SfM} & IMC2025 & HardMatch \\
    \cmidrule(lr){2-5} \cmidrule(lr){6-6} \cmidrule(lr){7-7}
    & $1^\circ$ & $3^\circ$ & $5^\circ$ & $10^\circ$ & score & mAA \\
    \midrule
    SuperPoint & \trd 68.9 & \trd 86.8 & \snd 92.3 & \snd 95.3 & \fst 66.6 & \fst 55.5 \\
    DaD & \snd 70.1 & \snd 86.9 & \trd 91.0 & \trd 94.5 & \snd 65.7 & \snd 55.3 \\
    RaCo & \fst 71.7 & \fst 88.1 & \fst 92.5 & \fst 95.4 & \fst 66.6 & \trd 54.9 \\
    \bottomrule
  \end{tabular}
\end{table}

\subsection{Keypoint detectors}
Since FlexGlue uses keypoints only as queries, it can be combined with any detector at test time without retraining.
\Cref{tab:detectors} evaluates the same model with 4096 SuperPoint~\citep{superpoint}, DaD~\citep{edstedt2025dad}, and RaCo~\citep{shenoi2026raco} keypoints, detected at 1024\,px long side, with a model input of 512\,px and sparse-to-sparse matching.
On ETH3D, we average the pose AUC over three COLMAP seeds and reject pairs with the covisibility threshold 0.9. On IMC2025, we report the ongoing leaderboard score.
All detectors perform similarly on two-view matching (HardMatch).
In multi-view reconstruction, RaCo yields the most accurate results on both ETH3D and IMC2025.

\subsection{Detailed results on LaMAR-DG}

 \begin{figure}[h]
      \centering
      \begin{subfigure}[b]{0.49\linewidth}
          \centering
          \includegraphics[width=\linewidth]{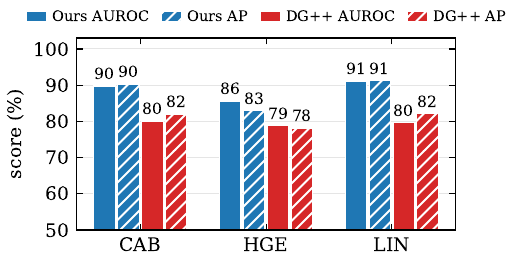}
          \caption{by scene}
          \label{fig:lamar-scene}
      \end{subfigure}\hfill
      \begin{subfigure}[b]{0.49\linewidth}
          \centering
          \includegraphics[width=\linewidth]{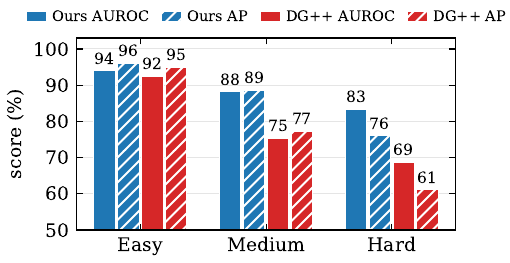}
          \caption{by difficulty}
          \label{fig:lamar-difficulty}
      \end{subfigure}
      \begin{subfigure}[b]{0.49\linewidth}
          \centering
          \includegraphics[width=\linewidth]{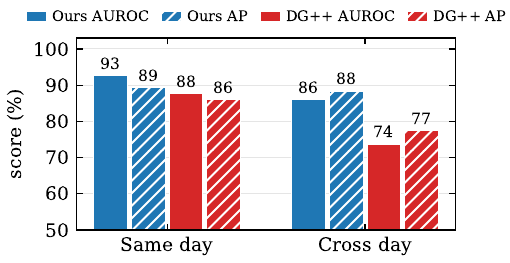}
          \caption{by time}
          \label{fig:lamar-day}
      \end{subfigure}\hfill
      \begin{subfigure}[b]{0.49\linewidth}
          \centering
          \includegraphics[width=\linewidth]{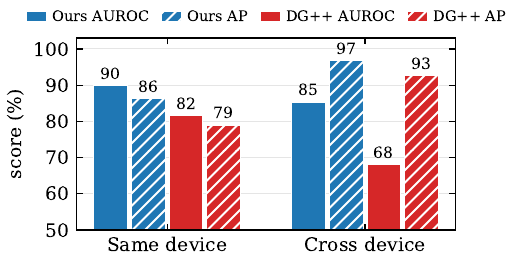}
          \caption{by device}
          \label{fig:lamar-device}
      \end{subfigure}
      \caption{\textbf{Detailed results on LaMAR-DG.} AUROC and AP of FlexGlue and DG++~\citep{xiangli2025doppelgangers++} per scene (a), difficulty (b), time (c), and device (d). Difficulty bins are tertiles of MegaLoc similarity per label, \ie, hard negatives are highly similar and hard positives dissimilar. FlexGlue remains robust on hard, cross-day, and cross-device pairs, where DG++ degrades strongly.}
      \label{fig:lamar-details}
  \end{figure}

In~\cref{fig:lamar-details}, we compare FlexGlue with DG++~\citep{xiangli2025doppelgangers++} on LaMAR-DG per building and group pairs by difficulty (MegaLoc similarity), time (same vs.\ cross day), and device. %
FlexGlue improves most on hard, cross-day, and cross-device pairs. Note that the high AP on cross-device pairs is caused by an imbalance of positives and negatives, an artefact of our hard negative mining (such pairs typically have low visual similarity).

\section{Benchmarks}

\subsection{Curation of LaMAR-DG}
\label{sec:supp_lamar}
We use the map sessions of the CAB, HGE, and LIN buildings, captured with HoloLens and iPhone devices.
We render depth for each view from the laser-scanned meshes and keep only views where at least 10 COLMAP points agree with the rendered depth within 10\%.
The overlap from view $i$ to view $j$ is the fraction of pixels of $i$ that project into $j$ with a depth consistent within 10\% of the rendered depth of $j$, which accounts for occlusions.
Positives are sampled uniformly among pairs with an overlap between 10\% and 80\% in both directions.
For negatives, we visit query views in farthest-point order over their camera centers and select the most similar non-overlapping view among the 50 nearest MegaLoc~\citep{Berton_2025_CVPR} neighbors, if its similarity exceeds 0.3.
This yields 500 positive and 500 negative pairs per building.
We group pairs into easy, medium, and hard tertiles of MegaLoc similarity per label, \ie, hard pairs are dissimilar positives and similar negatives.

\subsection{ViSym manual label cleaning}
\label{sec:supp_visym}
\begin{figure}[t]
  \centering
  \includegraphics[width=0.196\linewidth]{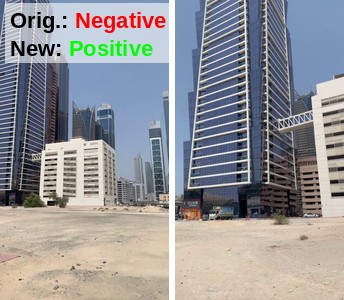}\hfill
  \includegraphics[width=0.196\linewidth]{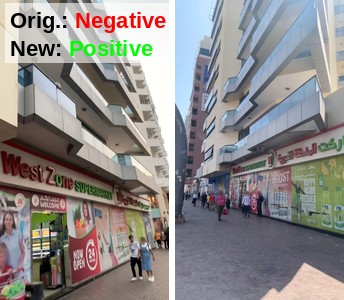}\hfill
  \includegraphics[width=0.196\linewidth]{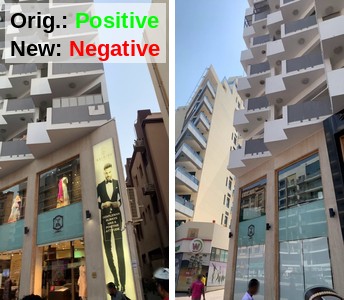}\hfill
  \includegraphics[width=0.196\linewidth]{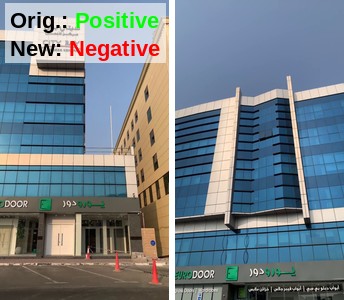}\hfill
  \includegraphics[width=0.196\linewidth]{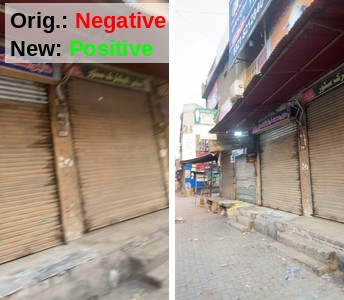}
  \caption{\textbf{Corrected ViSym test labels.} Each pair shows the original ViSym label (Orig.) and our label (New). The original labels miss covisible pairs under strong viewpoint changes and mark some pairs of distinct but similar facades as covisible.}
  \label{fig:visym_relabel}
\end{figure}

Initial training on ViSym~\citep{xiangli2025doppelgangers++} dropped our accuracy on LaMAR-DG, which is based on very accurate meshes and thus almost noise-free covisibility labels.
We observed the same effect in the official DG++ model,~\cref{tab:doppelgangers}: Trained on ViSym, its accuracy on LaMAR drops from 78.6\% to 73.3\% AUROC.
From visual inspection, we found that the GPS-derived labels of the ViSym dataset~\citep{xiangli2025doppelgangers++} are noisy and often mark covisible pairs as negatives (\cref{fig:visym_relabel}).
We therefore relabel all 3180 test pairs by hand, taking both the noisy GPS positions and the image content into account, and mark pairs as unsure when the decision is ambiguous, \eg, under dynamic occlusions or when only the background is covisible, and exclude those.
Most corrections turn negatives into positives (\cref{tab:visym_relabel}).
We then rebalance the test set with an equal number of positives and negatives per scene, at most 100 each, and evaluate all methods on this subset (ViSym-F).
\begin{table}[t]
    \caption{\textbf{ViSym test relabelling.} We manually clean GPS-annotated labels of ViSym by hand. Most corrections turn negatives into positives. The last row counts the pairs kept in our balanced subset.}
  \label{tab:visym_relabel}
  \centering
  \footnotesize
  \begin{tabular}{@{}lcccc@{}}
    \toprule
    Original label & Pairs & Unchanged & Flipped & Unsure \\
    \midrule
    Positive & 1590 & 1500 & 57 & 33 \\
    Negative & 1590 & 1154 & 418 & 18 \\
    \midrule
    Total & 3180 & 2654 & 475 & 51 \\
    \midrule
    Kept & 1184 & 946 & 238 & 0 \\
    \bottomrule
  \end{tabular}

\end{table}

\section{Implementation details}

\subsection{Training data}
\begin{table}[t]
  \caption{\textbf{Training datasets.} Number of training scenes and relative sampling weight of each dataset. Doppelgangers is only mixed in during fine-tuning with negatives.}
  \label{tab:datasets}
  \centering
  \footnotesize
  \begin{tabular}{@{}llrr@{}}
    \toprule
    Dataset & Type & \#Scenes & Weight \\
    \midrule
    MegaDepth~\citep{li2018megadepth} & outdoor, real & 152 & 0.17 \\
    AerialMegaDepth~\citep{aerialmegadepth} & aerial--ground, rendered & 113 & 0.17 \\
    BlendedMVS~\citep{yao2020blendedmvs} & objects \& scenes, rendered & 493 & 0.15 \\
    ScanNet++~\citep{yeshwanth2023scannet++} & indoor, real & 827 & 0.15 \\
    Hypersim~\citep{hypersim} & indoor, synthetic & 399 & 0.10 \\
    Map-free~\citep{mapfree} & outdoor, real & 460 & 0.10 \\
    DL3DV~\citep{dl3dv} & indoor \& outdoor, real & 9862 & 0.10 \\
    TartanAir~\citep{tartanair2020iros} & indoor \& outdoor, synthetic & 56 & 0.06 \\
    \midrule
    Doppelgangers~\citep{cai2023doppelgangers} & outdoor, real & 206 & 1.00 (stage 2) \\
    \bottomrule
  \end{tabular}
\end{table}

We list our full training data in~\cref{tab:datasets}.
We sample each dataset according to its weight and scenes uniformly within each dataset.
We sample pairs with dense visual overlap, normalized by area with depth coverage, in $[0.05, 0.8]$.
Throughout training, we apply light photometric augmentations, and in the final fine-tuning stage additionally continuous rotation and perspective augmentations.
In the final fine-tuning stage, half of the pairs come from Doppelgangers~\citep{cai2023doppelgangers} and half from the remaining datasets, of which half are turned into negatives by horizontally flipping one of the two images.
For the indoor variant in \cref{tab:ablation_dg}, we additionally mine pairs from the mapping sessions of four NaverLabs~\citep{lee2021naverlabs} floors, using every fourth frame and holding out a fifth floor for validation.
We mine positives and hard negatives with the same technique as for LaMAR-DG (\cref{sec:supp_lamar}).

\subsection{Supervision}
\label{sec:supp_supervision}
With 1.5-entmax, the cross-entropy is undefined for cells with zero probability.
We instead supervise the readout with the Fenchel--Young loss of $\alpha$-entmax~\citep{blondel2020learning} with $\alpha{=}1.5$:
\begin{align}
    \losscoarse = \langle \mathbf{P}_i - \mathbf{w}_i, \mathbf{z}_i \rangle + H_\alpha(\mathbf{P}_i) - H_\alpha(\mathbf{w}_i),
    \quad
    H_\alpha(\mathbf{P}) = \frac{1 - \sum_j P_j^\alpha}{\alpha(\alpha - 1)}
    \enspace,
\end{align}
where $\mathbf{z}_i$ are the logits with $z_{ij}{=}s_{ij}/\temperature$, $\mathbf{P}_i{=}\alpha\text{-entmax}(\mathbf{z}_i)$, $\mathbf{w}_i$ holds the bilinear weights of the ground-truth correspondence over its four surrounding cells and is zero elsewhere, and $H_\alpha$ is the Tsallis entropy~\citep{tsallis1988possible}.
Its gradient with respect to $\mathbf{z}_i$ is $\mathbf{P}_i - \mathbf{w}_i$, such that it remains finite when entmax assigns zero probability, and it is minimized at $\mathbf{P}_i{=}\mathbf{w}_i$.
Since the $3{\times}3$ window around any of the four target cells contains all of them, \cref{eq:readout} then recovers the ground-truth position exactly.

The refinement is supervised with the generalized Charbonnier loss~\citep{barron2019general,edstedt2023roma}:
\begin{align}
    \lossfine = \lossrobust\!\left(\norm{\offset^i - \left(\mathbf{p}^{\text{gt}}_i - \operatorname{sg}(\corrab^i)\right)}_2\right),
    \quad
    \lossrobust(x) = \frac{|\alpha - 2|}{\alpha} \left( \left( \frac{(x/c)^2}{|\alpha - 2|} + 1\right)^{\frac{\alpha}{2}} -1 \right)
    \enspace,
\end{align}
with $\alpha{=}0.5$ and $c{=}0.24$\,px, where $\mathbf{p}^{\text{gt}}_i$ is the ground-truth correspondence and $\operatorname{sg}$ the stop-gradient.
All heads are instantiated separately for each layer, and we average the losses uniformly over layers.

\subsection{Matching modes}
\label{sec:supp_modes}
FlexGlue supports three matching modes, which differ only in the queries and in how correspondences are converted into matches.
In all modes, we discard correspondences with a confidence $\confid^i$ below a threshold, 0.1 by default, or with a cycle error above 3\,px, \ie, if querying FlexGlue at $\corrab^i$ in the reverse direction does not land within 3\,px of $\kpt_a^i$.

\begin{enumerate}[label=(\roman*), leftmargin=*, topsep=2pt, itemsep=1pt]
    \item \textbf{Sparse-to-dense:} each query keypoint is matched to its predicted correspondence. With keypoints in both images, we concatenate the correspondences of both directions.
    \item \textbf{Sparse:} downstream applications such as SfM require point-to-point assignments between keypoints. We compute the symmetric distance between the keypoints and the predicted correspondences of both directions,
    \begin{align}
        D^{ij} = \tfrac{1}{2}\left(\lVert\corrab^i - \kpt_b^j\rVert_2 + \lVert\kpt_a^i - \corrba^j\rVert_2\right),
    \end{align}
    and select pairs of keypoints that are mutual nearest neighbors in $D$ with $D^{ij} < 3$\,px and $\confid_a^i \confid_b^j$ above the threshold.
    \item \textbf{Dense:} without keypoints, we sample up to 5k correspondences from dense predictions in both images, weighted by their confidence~\citep{edstedt2026romav2}.
\end{enumerate}

\section{Evaluation details}

\subsection{Relative Pose estimation}

We follow the evaluation protocols of the respective baselines.
Images in ScanNet-1500 are resized to $640{\times}480$\,px, following standard practice.
We estimate the essential matrix with the RANSAC of OpenCV and an inlier threshold of 0.5\,px for all methods.
We compute the pose error as the maximum of the angular rotation and translation errors and report the area under the cumulative recall curve (AUC) at $5\degree / 10\degree / 20\degree$.
On HardMatch, we estimate the fundamental matrix $F$ from correspondences, compute the epipolar error of the GT correspondences, and report the mean average accuracy (mAA) at 10\,px.

\paragraph{Baselines:}
For sparse matching, all methods detect up to 4096 keypoints, and resize images to 1600px for SuperPoint~\citep{superpoint} and ALIKED~\citep{Zhao2023ALIKED}.
Semi-dense and dense matchers differ in their evaluation protocol, and typically estimate many more correspondences.
We evaluate MASt3R~\citep{leroy2024grounding} (symmetric), EDM~\citep{li2025edm}, RoMa~\citep{edstedt2023roma}, UFM~\citep{zhang2025ufm} and RoMa v2~\citep{edstedt2026romav2}.
In the sparse-to-dense category, we benchmark CAPS~\citep{wang2020learning} and S2DNet~\citep{germain2020s2dnet}, two monocular baselines which perform an exhaustive search over dense features, and RCM~\citep{lu2024raising}, a deep sparse-to-dense matcher. We further pair the SotA dense matcher RoMav2~\citep{edstedt2026romav2} with SuperPoint keypoints, where we interpolate the dense correspondence maps at keypoint locations. In this setting, we detect only 2048 keypoints to remain roughly comparable to sparse methods, which are limited by keypoint repeatability.

\subsection{Multi-view estimation}
\paragraph{Baselines:}
On ETH3D, we report results with 2048 keypoints matched with LightGlue~\citep{lindenberger2023lightglue}, RoMav2~\citep{edstedt2026romav2} and LoMa-B~\citep{nordstrom2026loma}. We also show the symmetry-aware baseline MASt3R-SfM~\citep{duisterhof2025mast3r} with DG++~\citep{xiangli2025doppelgangers++} visual disambiguation, following the original implementation.
We further strengthen baselines with Doppelgangers++ disambiguation.
We run COLMAP~\citep{schoenberger2016sfm} SfM via hloc~\citep{sarlin2019coarse} with default settings and exhaustive pair selection, unless otherwise stated.

\subsection{Visual disambiguation}

\paragraph{Setup:}
Visual disambiguation is closely related to view graph filtering in SfM, where ambiguous correspondences on non-covisible image pairs can lead to catastrophic failures, such as scene collapses.
We consider three benchmarks: Doppelgangers~\citep{cai2023doppelgangers} contains 4659 challenging phototourism image pairs that often require analyzing fine-grained structural details. ViSym~\citep{xiangli2025doppelgangers++} is a collection of 3180 phone-captured street scene pairs from around the globe, which we manually clean (\cref{sec:supp_visym}). Our LaMAR-DG benchmark contains 3000 indoor and outdoor pairs (\cref{sec:supp_lamar}).

\paragraph{Baselines:} We compare against the foundational works in visual disambiguation: Doppelgangers (DG)~\citep{cai2023doppelgangers}, which uses a CNN encoder conditioned on local feature matches, and DG++~\citep{xiangli2025doppelgangers++}, which adds a lightweight decoder on top of the frozen dense matcher MASt3R~\citep{leroy2024grounding}.
We further report a variant of our method trained without labelled doppelganger pairs, using only flip augmentations and non-overlapping pairs from our matching datasets (``matching'' in \cref{tab:doppelgangers}).

\subsection{Visual Localization}
\paragraph{Baselines:}
On InLoc,~\cref{tab:inloc}, we compare against the sparse matchers SP+SG~\citep{sarlin2020superglue}, SP+LG~\citep{lindenberger2023lightglue} and LoMa~\citep{nordstrom2026loma} as well as the semi-dense and dense baselines EDM~\citep{li2025edm}, MASt3R~\citep{leroy2024grounding}, RoMa~\citep{edstedt2023roma} and RoMav2~\citep{edstedt2026romav2}.
We report the percentage of correctly localized queries at $(0.25\text{m}, 10\degree)$ / $(0.5\text{m}, 10\degree)$ / $(1.0\text{m}, 10\degree)$ and list the official numbers from the respective papers. We follow the hloc~\citep{sarlin2019coarse} evaluation protocol. For each query, we retrieve the top-40 database images, establish 2D-2D correspondences per pair, lift them to 2D-3D using the database depth, and estimate the absolute pose with PnP. We use grid keypoints (stride 8) as queries for our method.

\section{Limitations}
\begin{itemize}
    \item \textbf{Error correlation.}
    Experimentally, we observe that the errors of different queries are correlated because of the forced, learned interpolation between cells on the coarse grid.
    Despite a high precision, this can reduce the accuracy of relative pose estimation, which assumes the errors of points to be independent.
    In particular, most errors are also along the epipolar line (61.0 PCK@1px vs.\ 88.4\% epipolar precision @1px on the validation set), suggesting that FlexGlue learns some kind of epipolar geometry, which can be noisy.
    \item \textbf{Fine accuracy.} While our model achieves high accuracy and highest robustness, RoMav2~\citep{edstedt2026romav2} yields even more accurate matches on easy pairs because of its dense refiner. We believe that a stronger refinement model or a dedicated sparse-to-dense refiner with finer features could improve here.
    \item \textbf{Cross-Attention spread is local in later layers.} Empirically, we found that the coarse attention is often not improving much in later layers, and that the attention spread in cross-attention is very local deeper in the network.
    \item \textbf{Accuracy drops on depth discontinuities.} The network is forced to interpolate, and around depth discontinuities (where depth jumps) the accuracy drops (from $>$90\% PCK@5px to around $75\%$ PCK@5px). The cause of this is a bimodal distribution (landing on the wrong side of the discontinuity), and our interpolation of coarse features on the query side sometimes causes a correspondence to land on the wrong side. The proposed windowed readout already partially mitigates this issue on the target side, but incorporating finer features could potentially improve even further.

\end{itemize}
\section{Qualitative results}

\begin{figure*}[t!]
  \centering
  \begin{minipage}[t]{0.333\linewidth}\centering\small sparse-to-sparse\end{minipage}%
  \begin{minipage}[t]{0.333\linewidth}\centering\small matched keypoints\end{minipage}%
  \begin{minipage}[t]{0.333\linewidth}\centering\small sparse-to-dense\end{minipage}\\[2pt]
  \includegraphics[width=\linewidth]{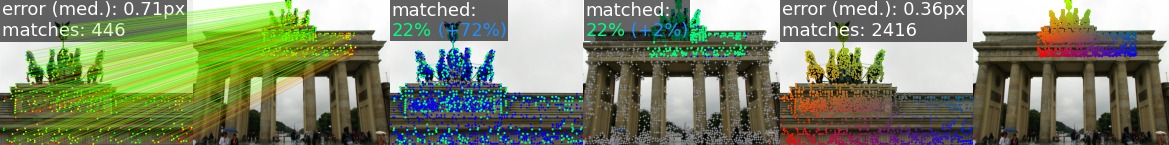}
  \includegraphics[width=\linewidth]{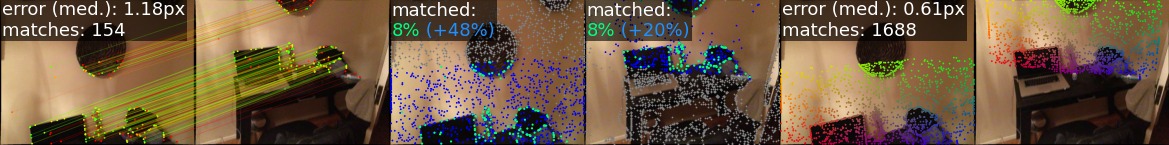}
  \caption{Comparison of matching modes. \textbf{Left:} Point-to-point correspondences estimated by our model, colored by epipolar error (green $< 1$\,px). \textbf{Center:} Keypoints colored by being matchable sparse-to-sparse (green), and keypoints only matchable sparse-to-dense (blue), highlighting the repeatability issue. \textbf{Right:} Sparse-to-dense correspondences. Especially under large scale changes and on low-texture areas, sparse-to-dense matching yields many more, accurate correspondences.}
  \label{fig:s2svss2d}
\end{figure*}

\begin{figure*}[p]
  \centering
  \newcommand{\pcarow}[2]{%
    \includegraphics[width=0.245\linewidth]{figures/pca/#1_views.jpg}\hfill
    \includegraphics[width=0.245\linewidth]{figures/pca/#1_pca.png}\hfill
    \includegraphics[width=0.245\linewidth]{figures/pca/#2_views.jpg}\hfill
    \includegraphics[width=0.245\linewidth]{figures/pca/#2_pca.png}\\[2pt]}
  \pcarow{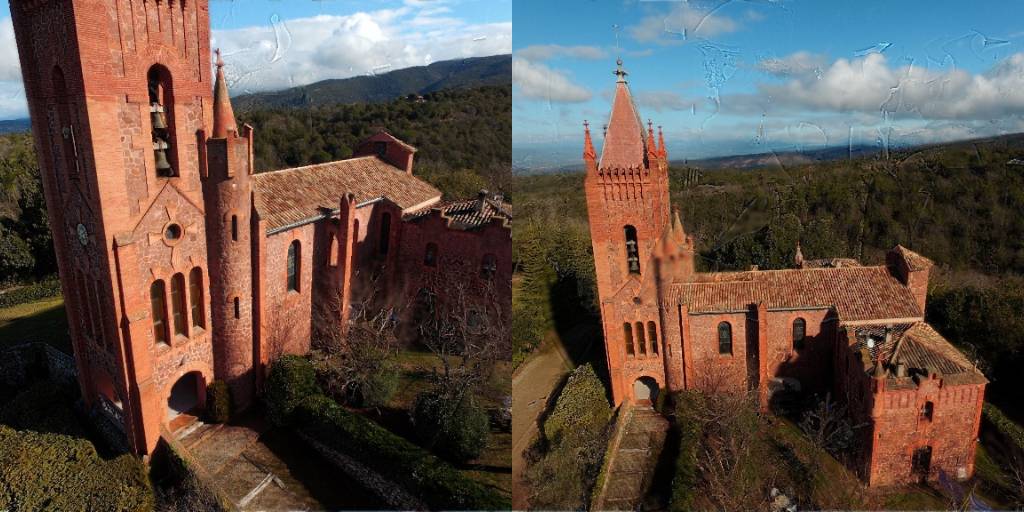}{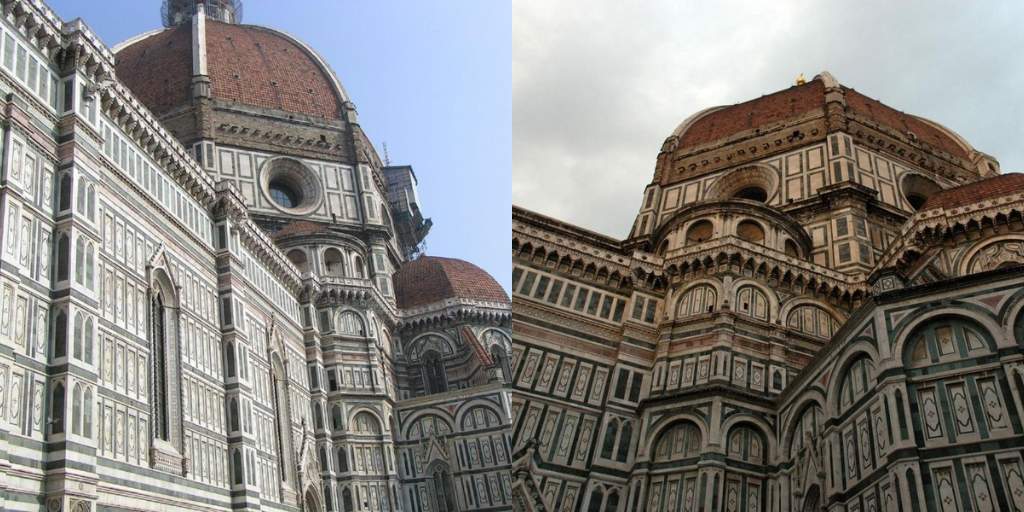}
  \pcarow{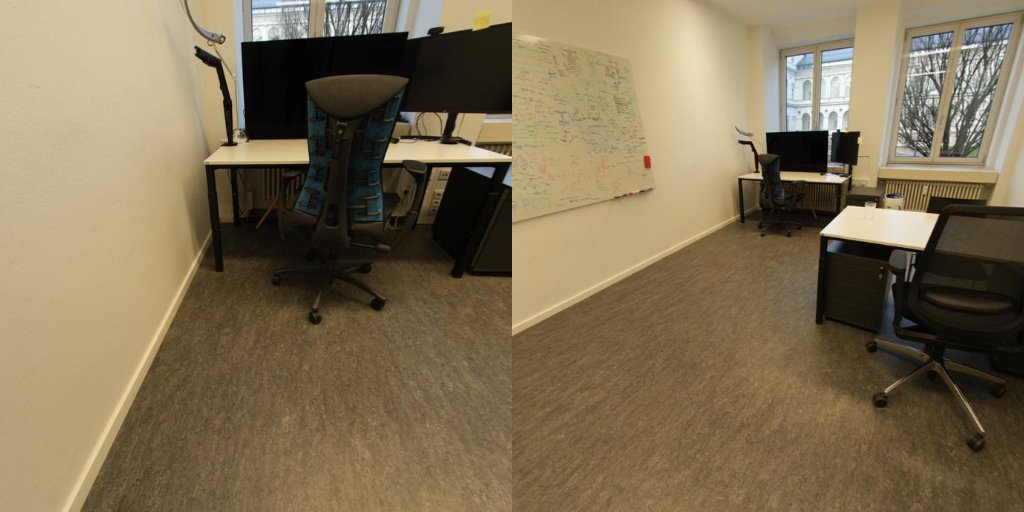}{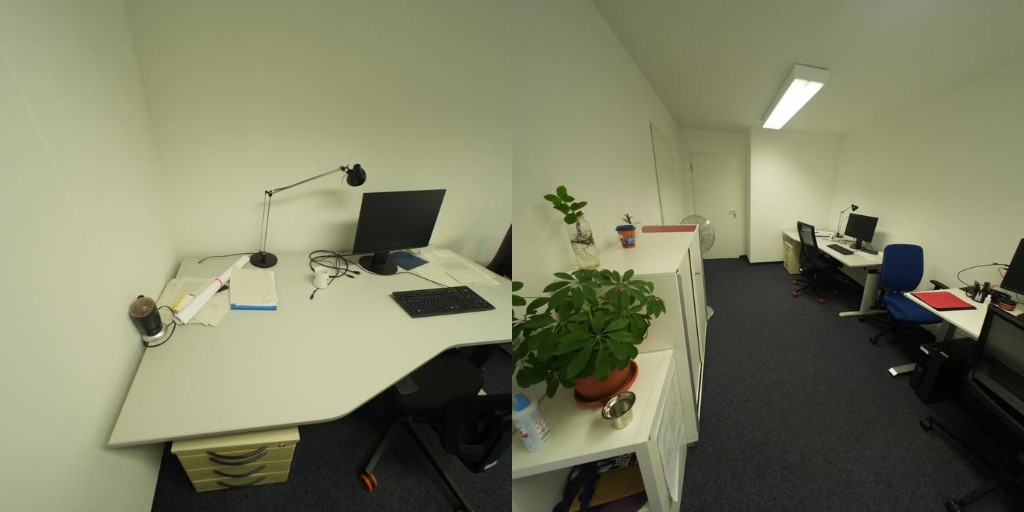}
  \pcarow{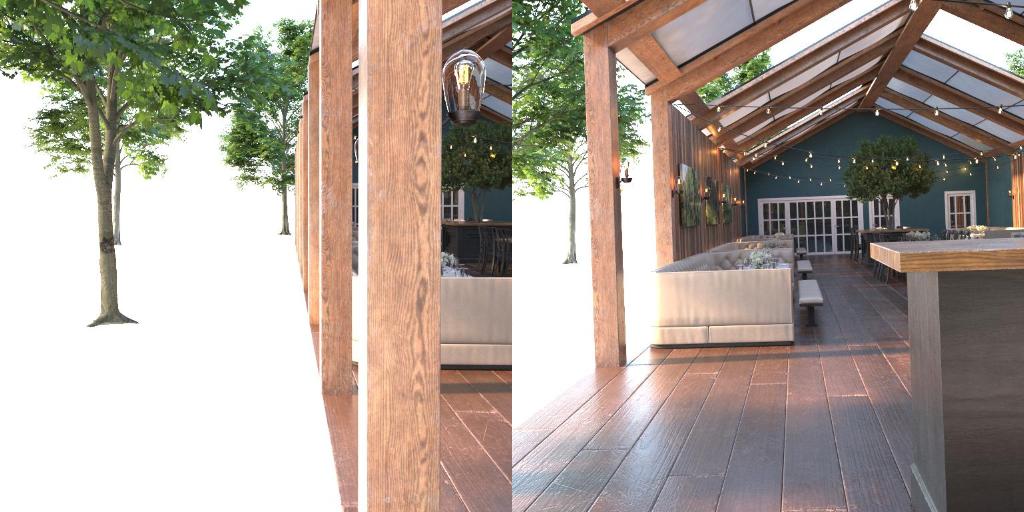}{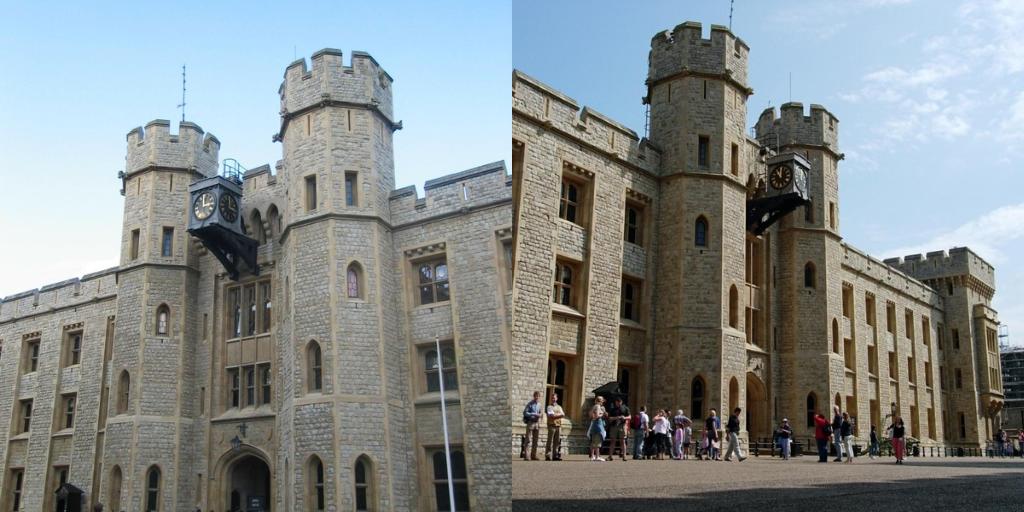}
  \pcarow{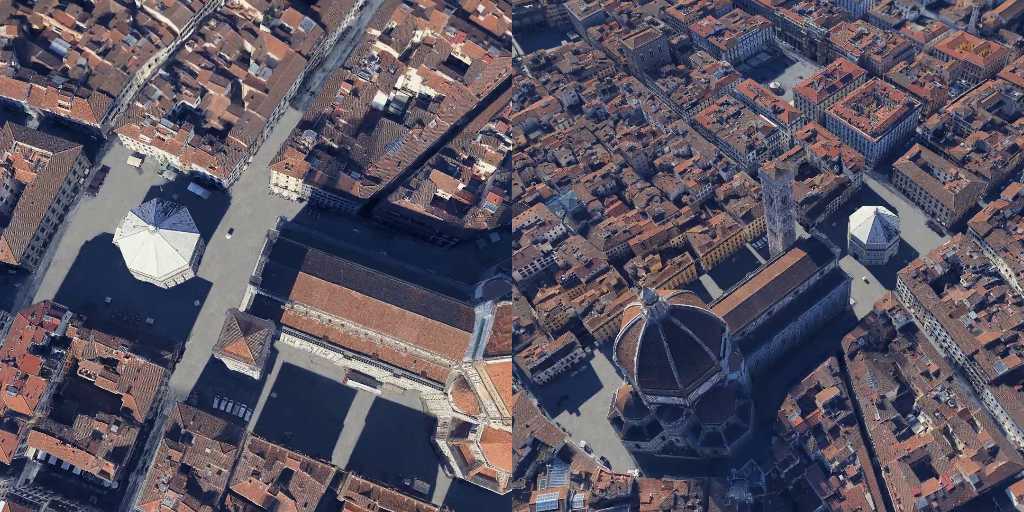}{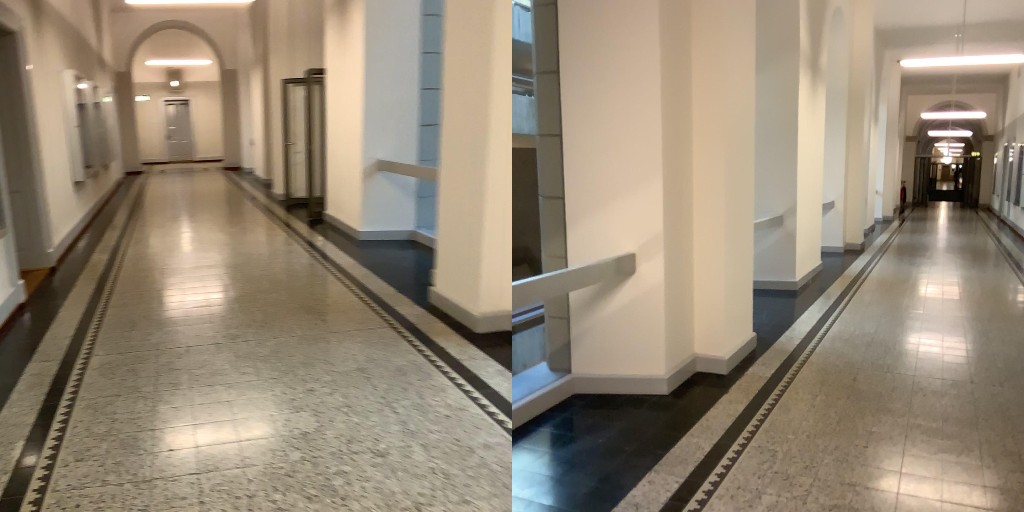}
  \pcarow{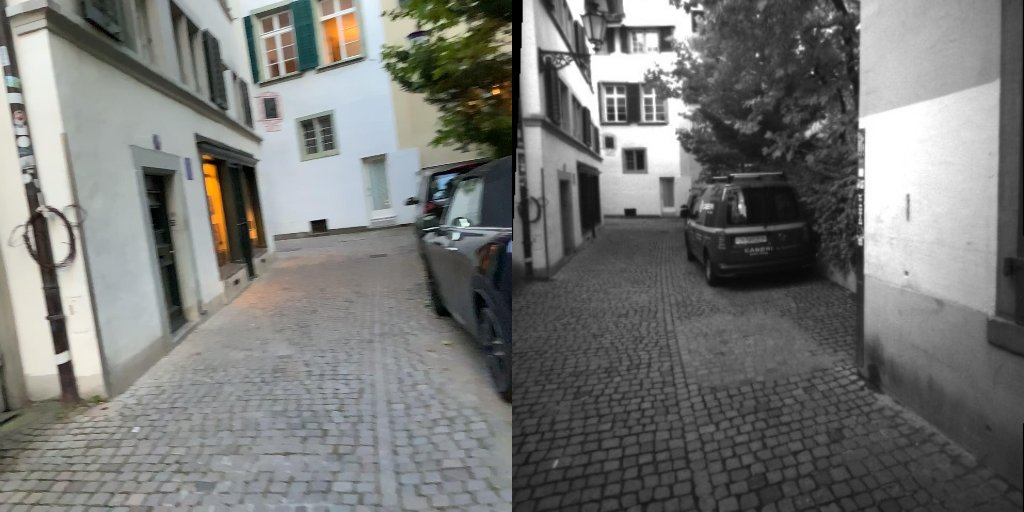}{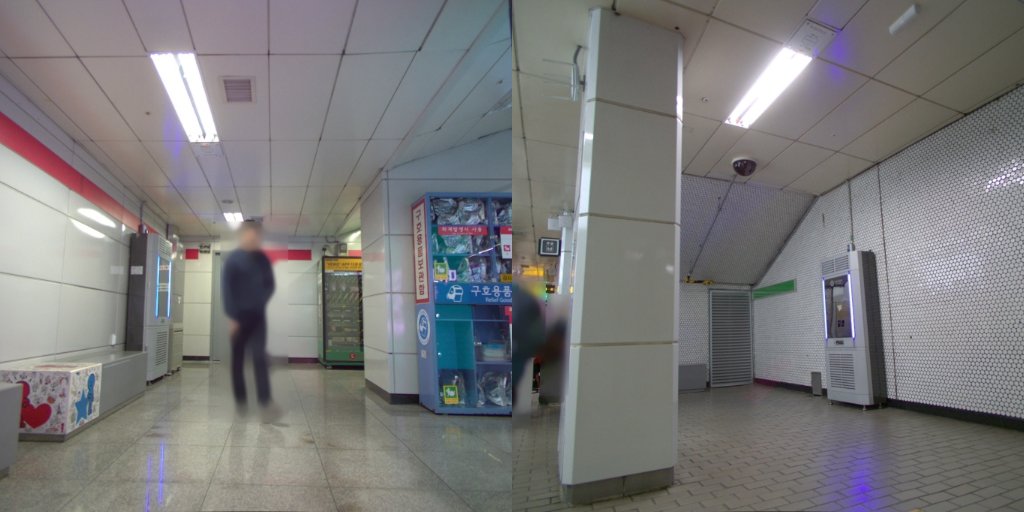}
  \pcarow{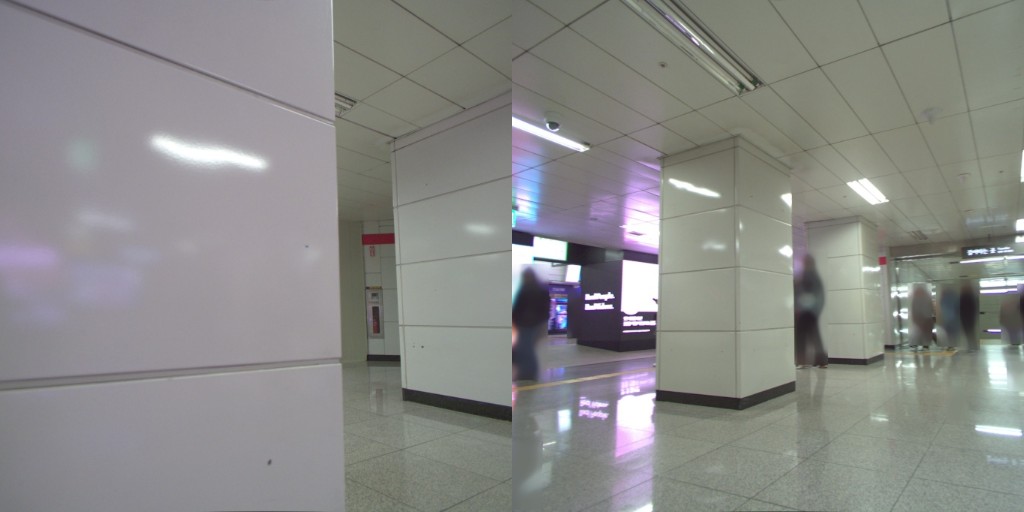}{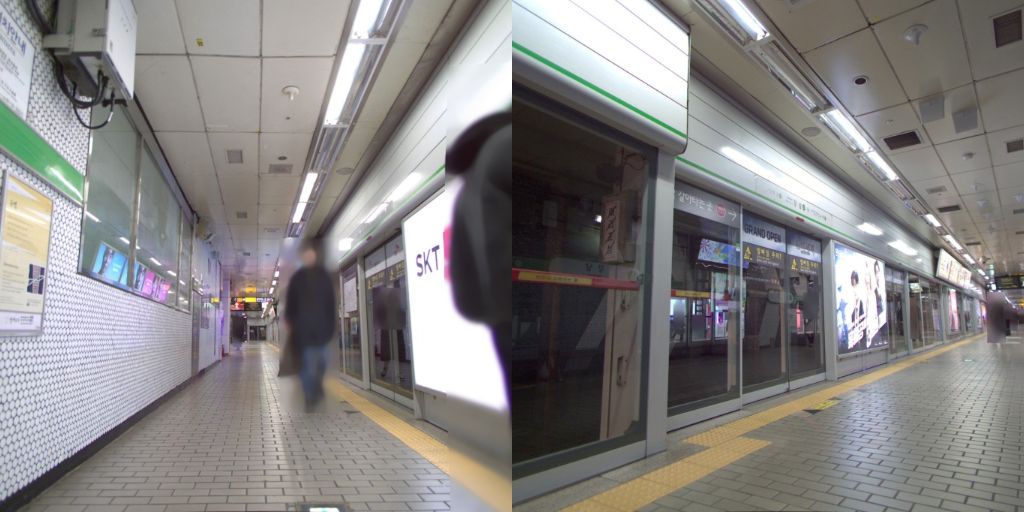}
  \pcarow{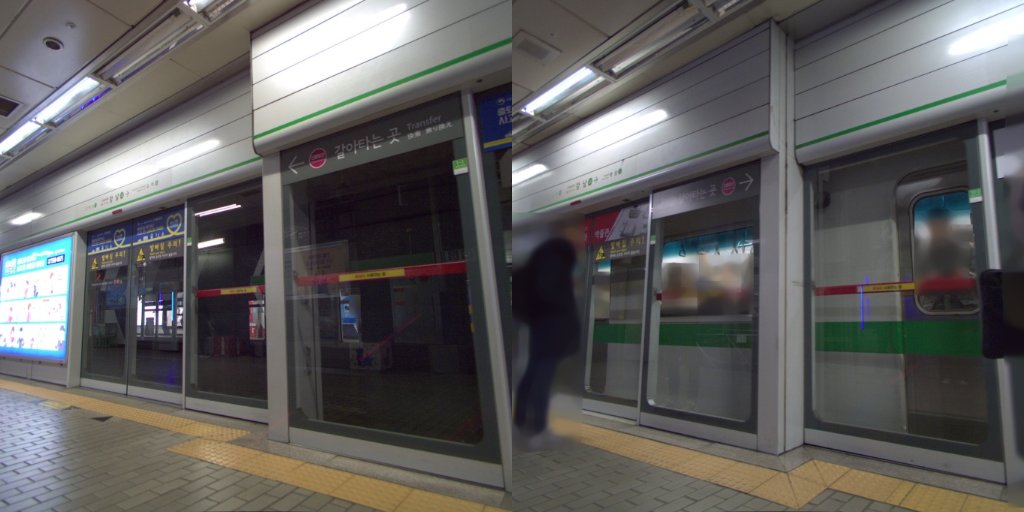}{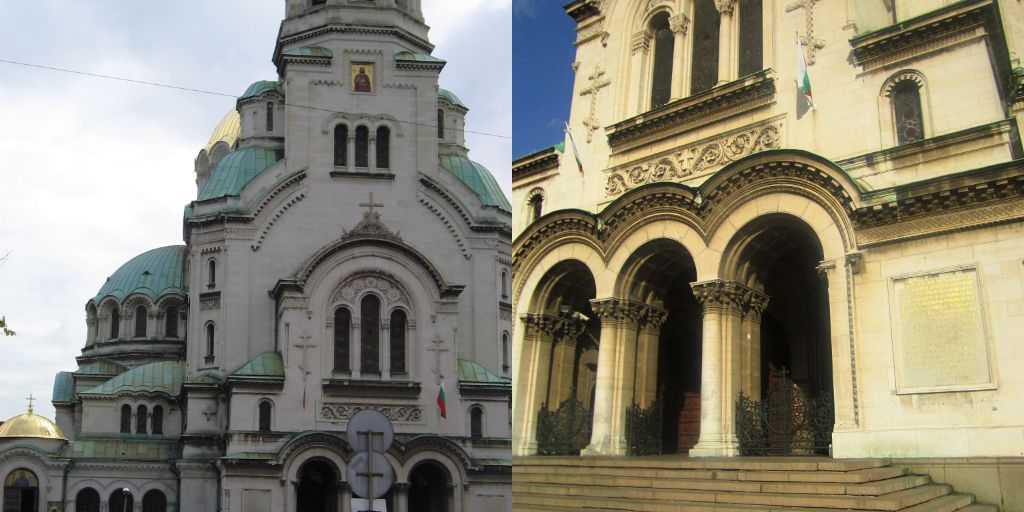}
  \pcarow{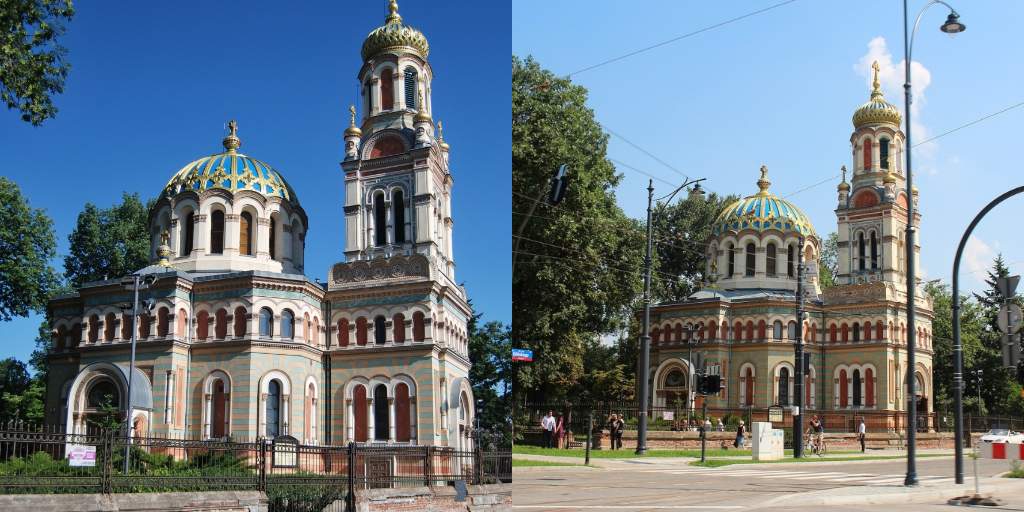}{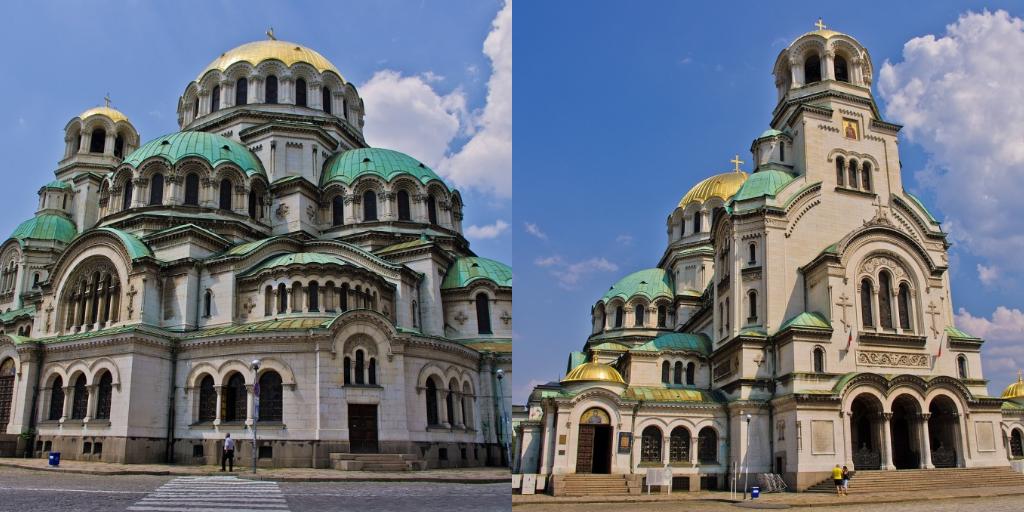}
  \pcarow{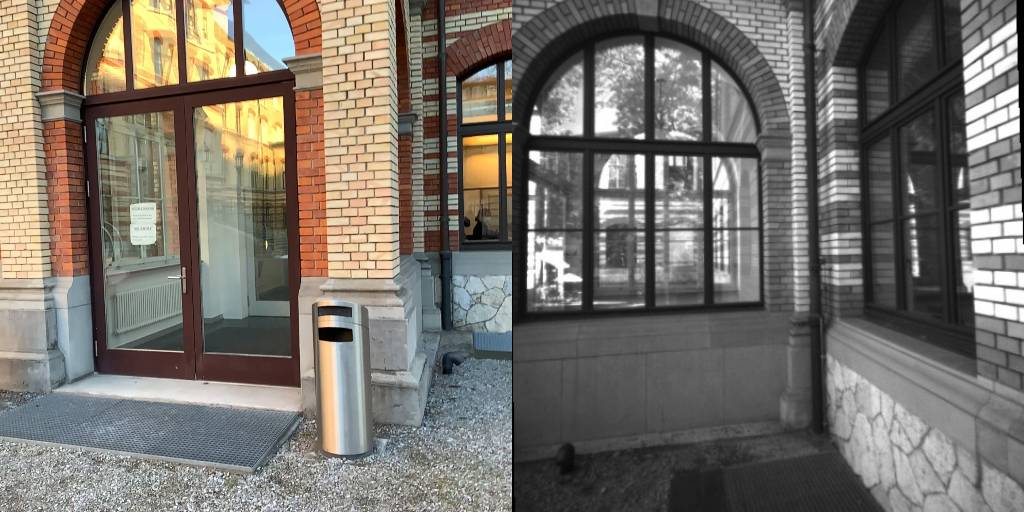}{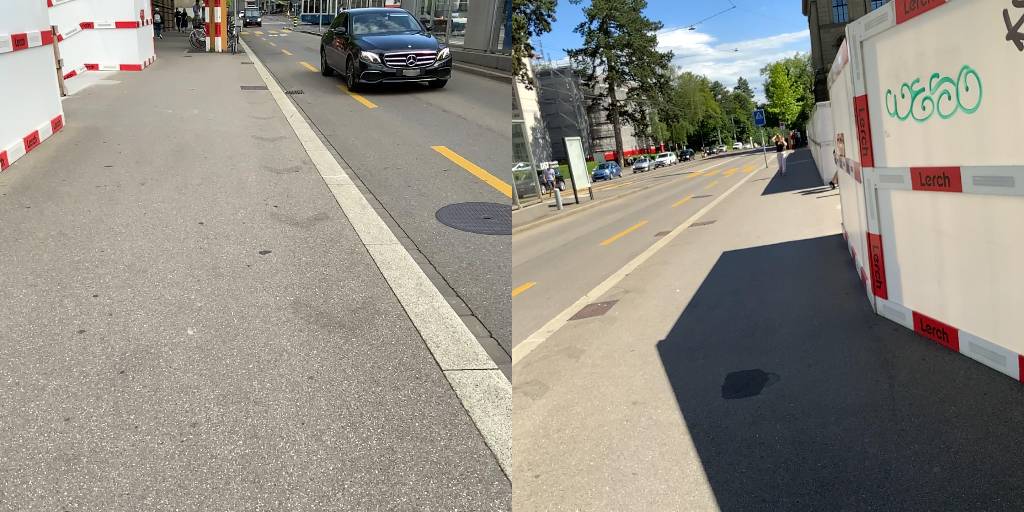}
  \pcarow{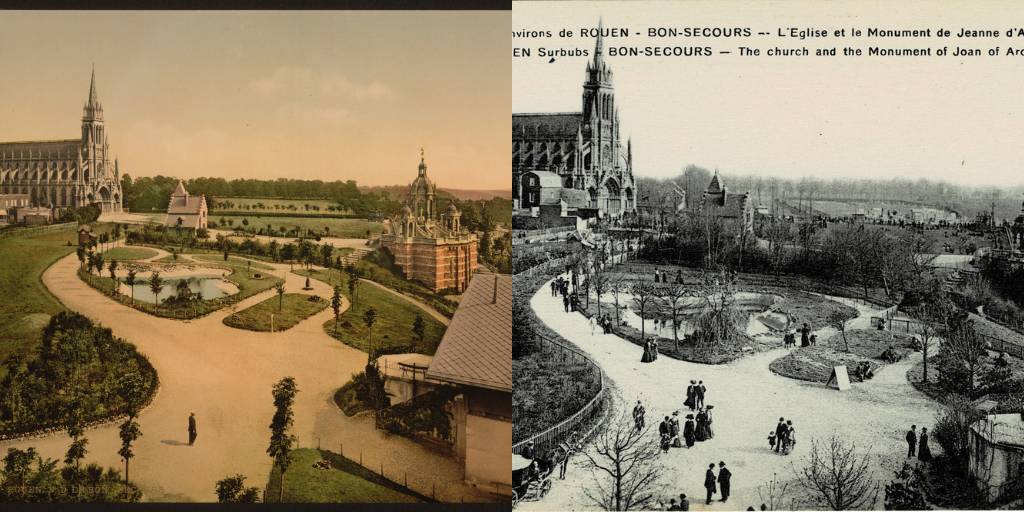}{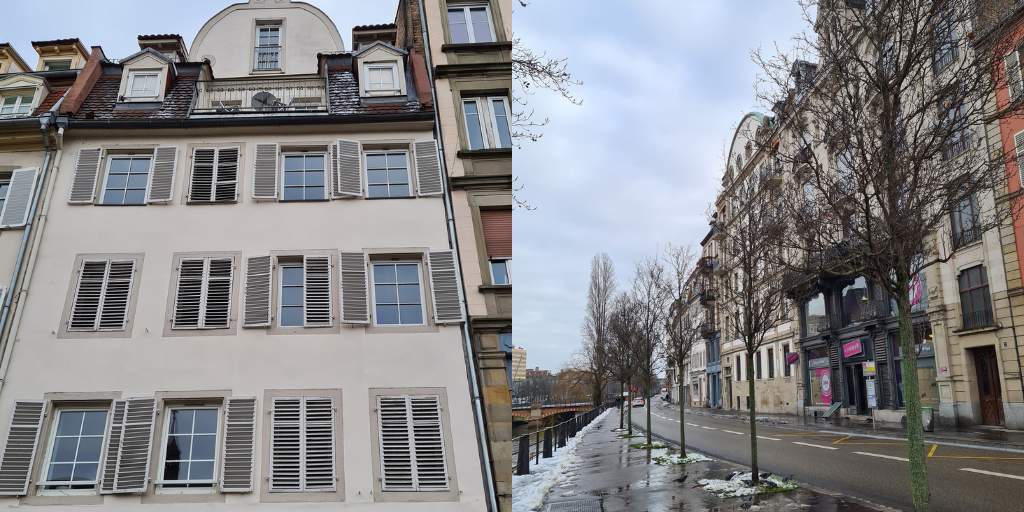}
  \pcarow{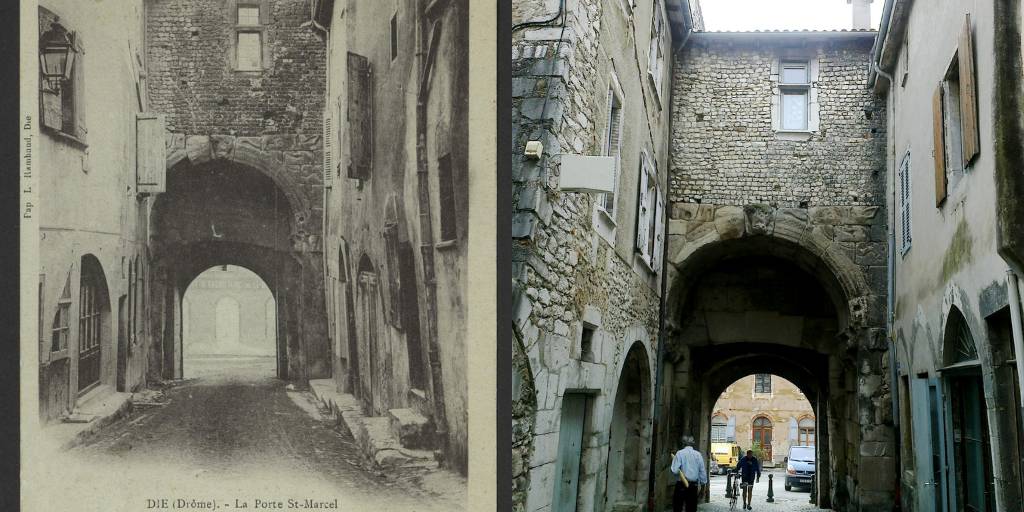}{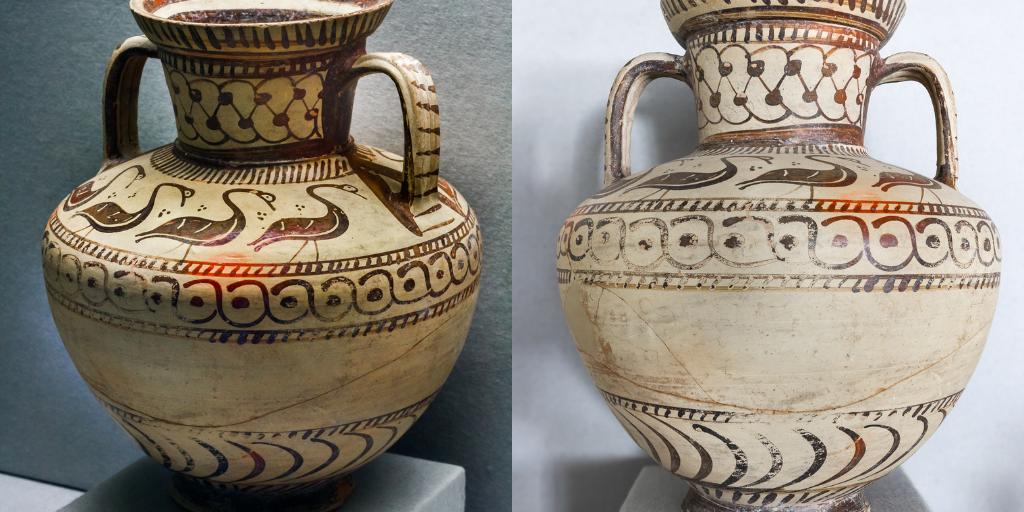}
  \pcarow{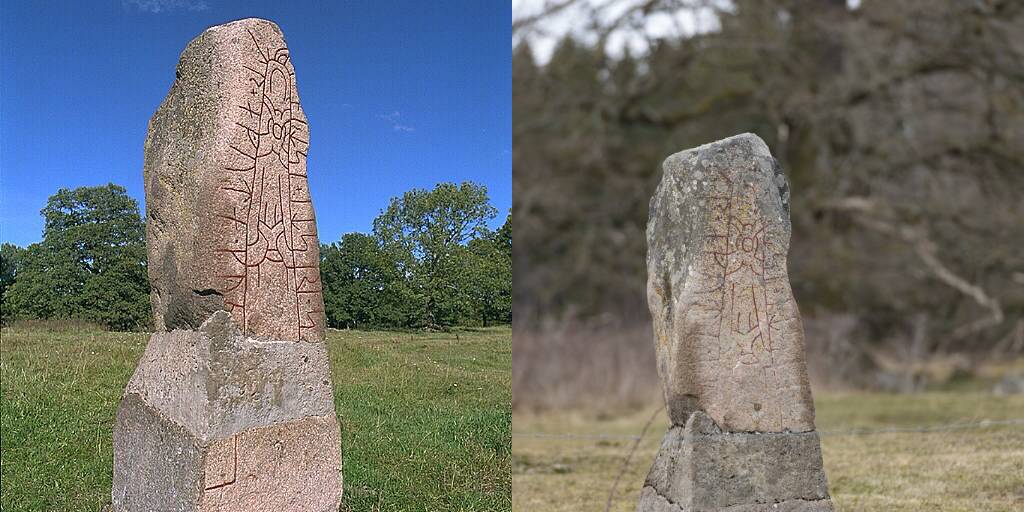}{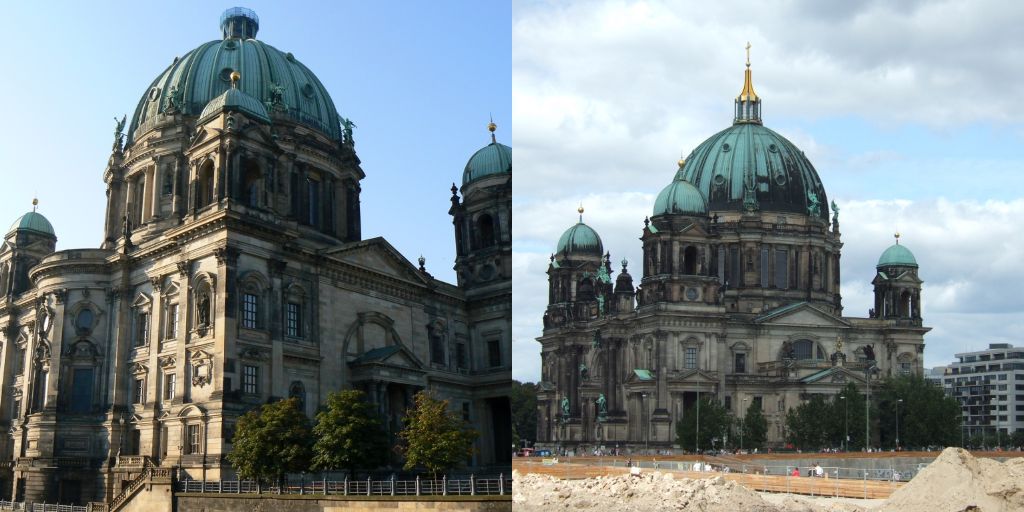}
  \caption{\textbf{PCA of the aggregated tokens.} For each image pair, we show the two views and the first three principal components of their tokens at 9 layers as RGB. Our method learns structural priors such as horizontal and vertical lines, perspective, and planarity, and clearly highlights covisible areas.}
  \label{fig:pca}
\end{figure*}

\begin{figure*}[t]
  \centering
  \setlength{\fboxsep}{0pt}
  \fbox{\includegraphics[height=0.29\linewidth]{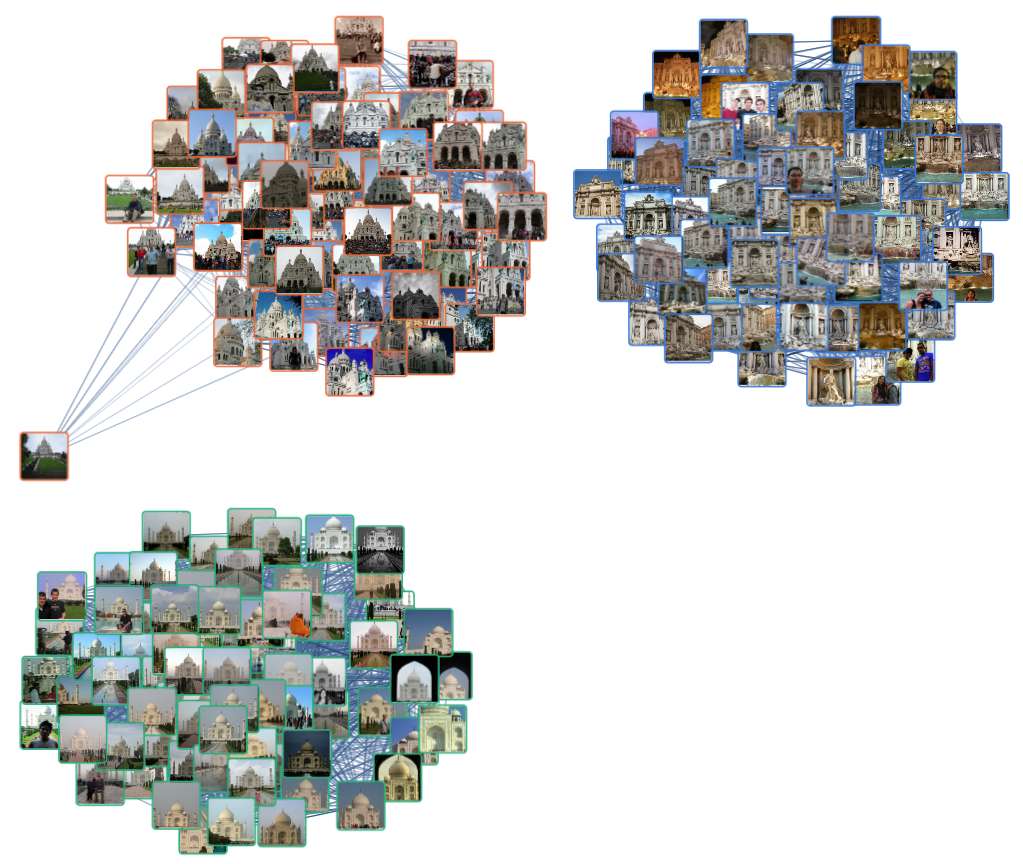}}\hfill
  \fbox{\includegraphics[height=0.29\linewidth]{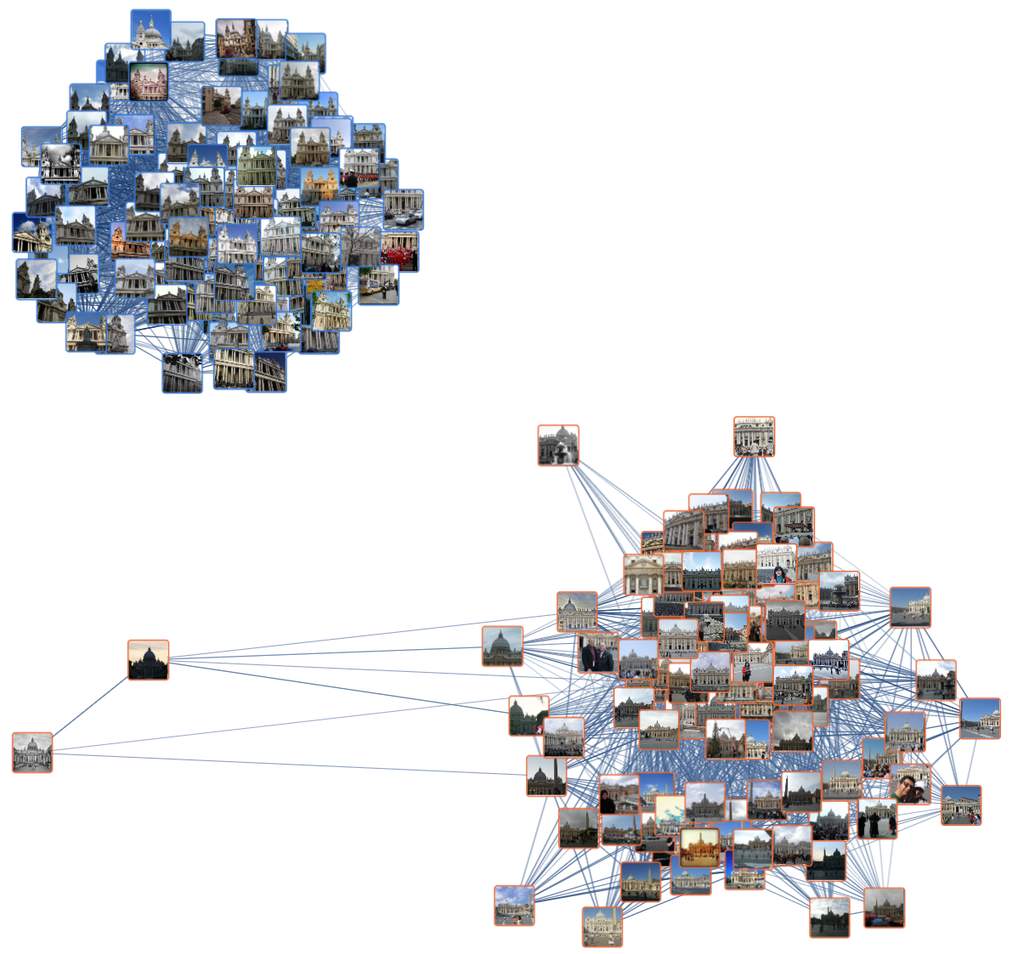}}\hfill
  \fbox{\includegraphics[height=0.29\linewidth]{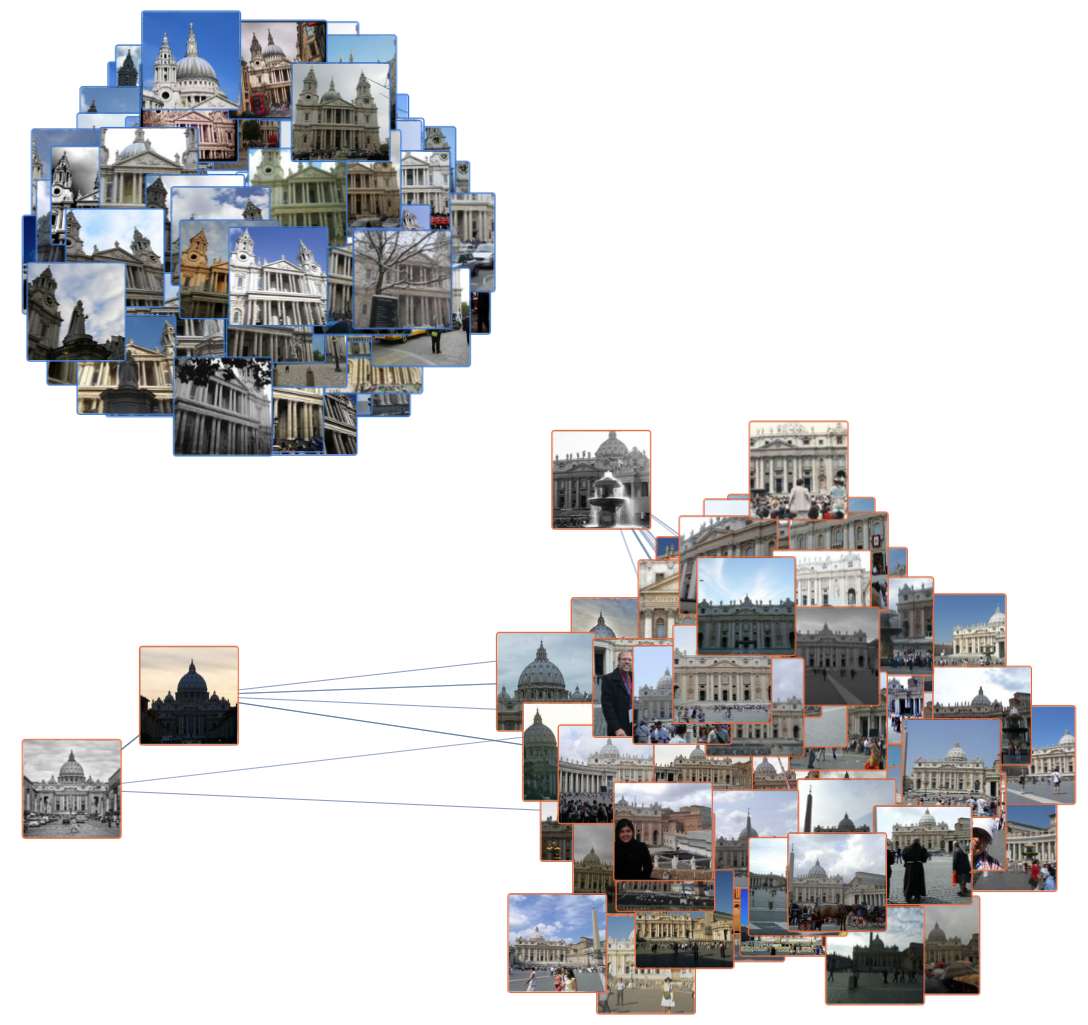}}\\[4pt]
  \fbox{\includegraphics[width=\dimexpr\linewidth-2\fboxrule]{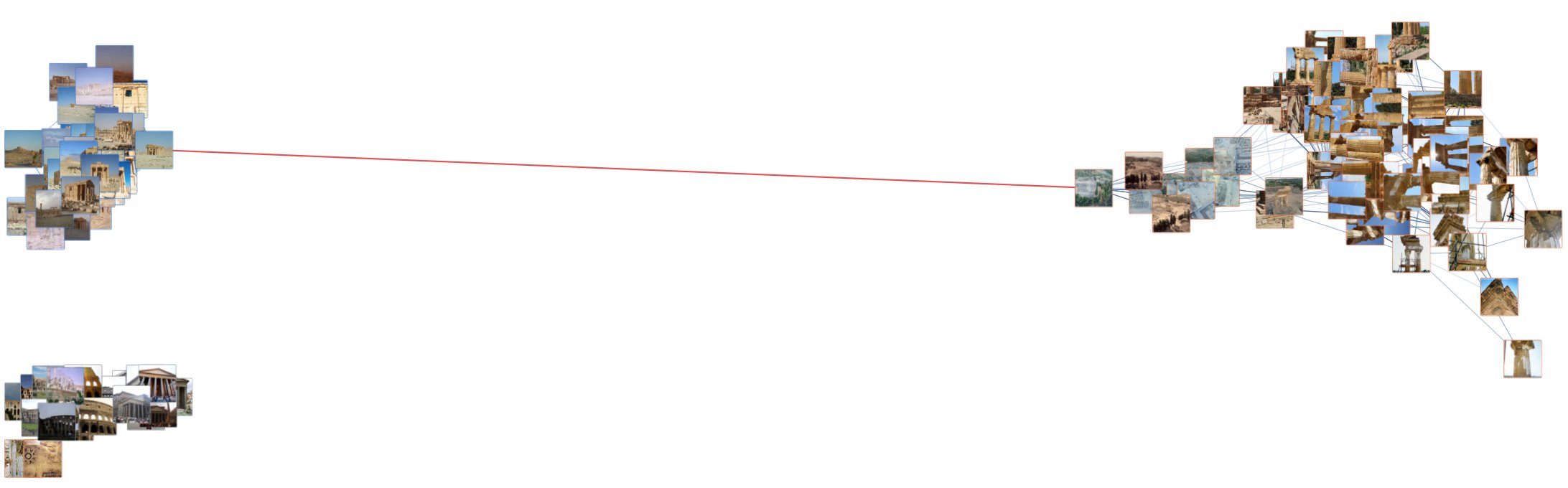}}
  \caption{\textbf{Scene clustering on IMC2025.} Image graphs recovered by our method with visual disambiguation only. Colors denote the predicted clusters and edges the retained image pairs. Top: landmarks are separated into clean clusters. Bottom: a single wrong pair (red) passes our filter.}
  \label{fig:imc_clusters}
\end{figure*}

\begin{figure*}[t]
  \centering
  {\footnotesize
  \makebox[0.333\linewidth]{Coarse readout}%
  \makebox[0.333\linewidth]{+ our refinement}%
  \makebox[0.333\linewidth]{+ RoMav2 refinement}}\\[1pt]
  \includegraphics[width=\linewidth]{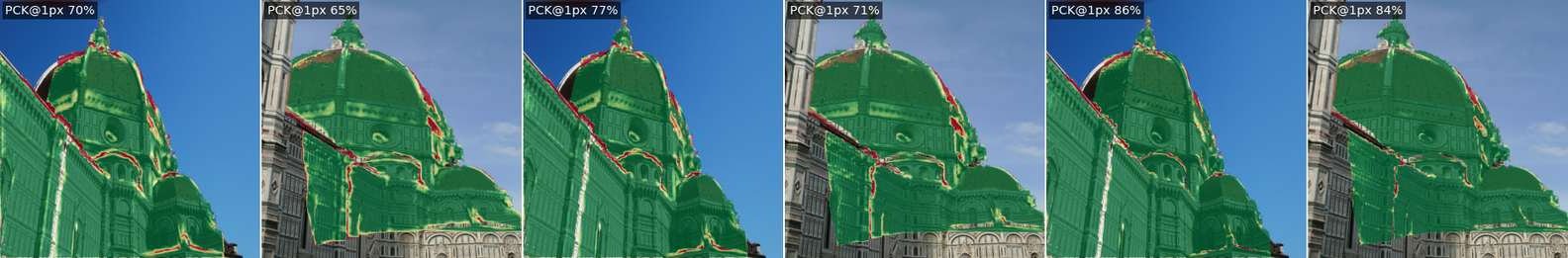}\\[2pt]
  \includegraphics[width=\linewidth]{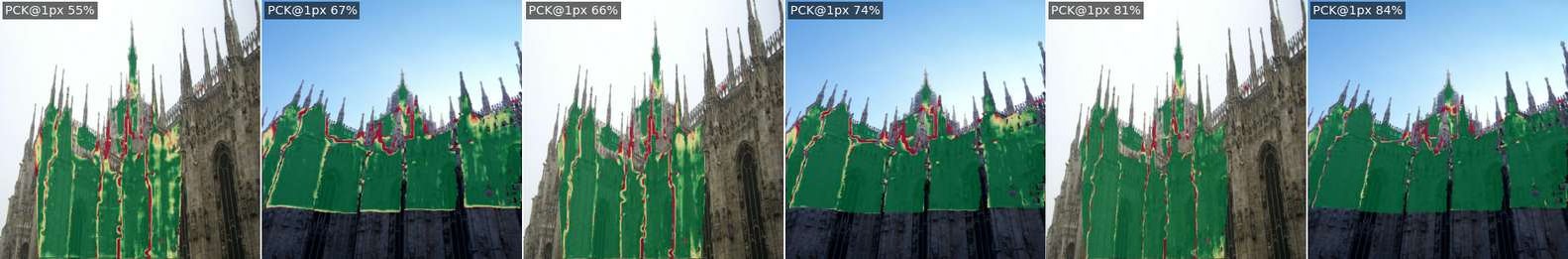}\\[2pt]
  \includegraphics[width=\linewidth]{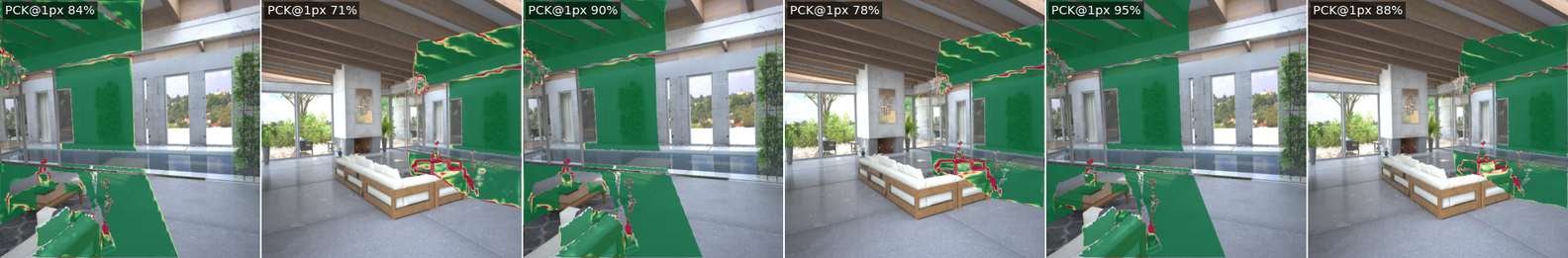}
  \caption{\textbf{Dense matching.} Dense correspondences of FlexGlue, colored by their error from green (${<}1$\,px) to red (${\geq}5$\,px), with the PCK@1px per image, from the coarse readout, with our refinement, and with the RoMav2~\citep{edstedt2026romav2} refiner. The coarse matching has weaknesses near depth discontinuities, which the refiners address.}
  \label{fig:dense_matches}
\end{figure*}

\begin{figure*}[t]
  \centering
  {\footnotesize
  \makebox[0.333\linewidth]{3 layers}%
  \makebox[0.333\linewidth]{6 layers}%
  \makebox[0.333\linewidth]{9 layers}}\\[1pt]
  \includegraphics[width=\linewidth]{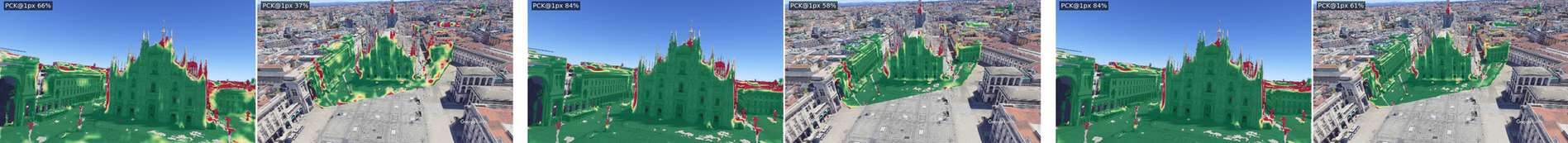}\\[2pt]
  \includegraphics[width=\linewidth]{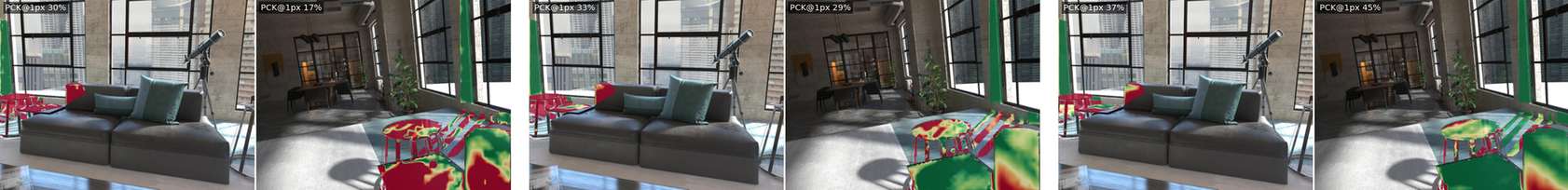}\\[2pt]
  \includegraphics[width=\linewidth]{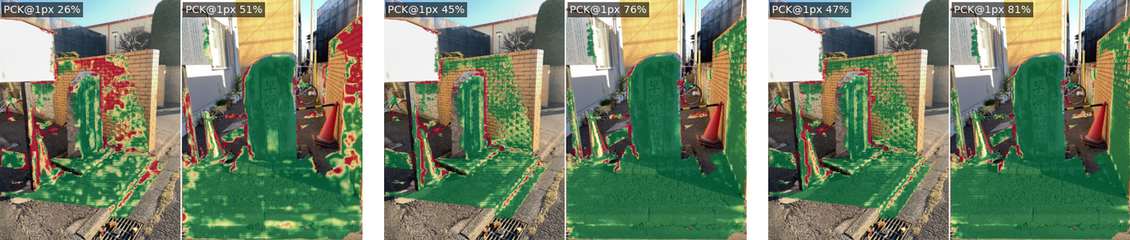}\\[2pt]
  \includegraphics[width=\linewidth]{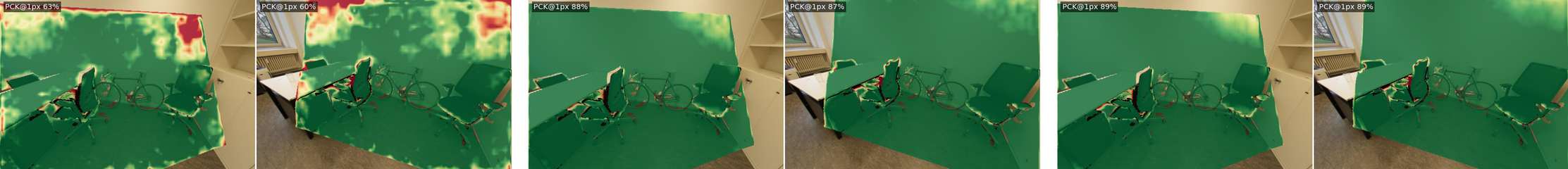}\\[2pt]
  \includegraphics[width=\linewidth]{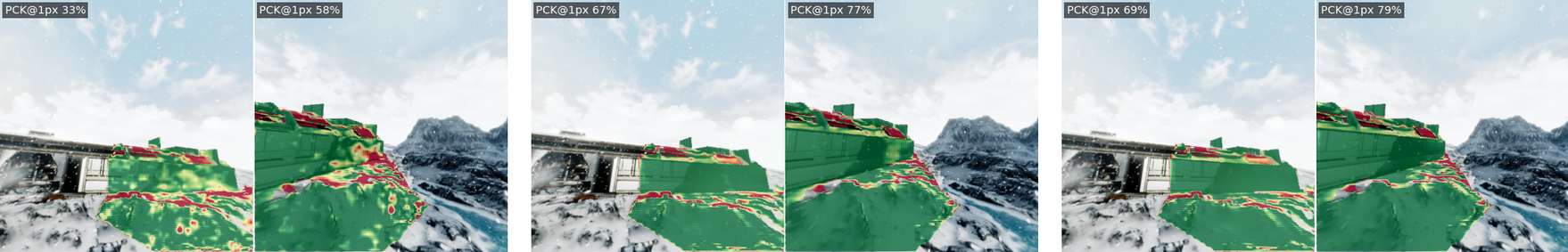}
  \caption{\textbf{Dense matching across layers.} Unfiltered dense correspondences of FlexGlue after 3, 6, and 9 layers, colored by their error from green (${<}1$\,px) to red (${\geq}5$\,px), with the PCK@1px per image. Despite being supervised only at sparse queries, FlexGlue yields fairly accurate dense correspondences. Deeper layers mostly refine matches and fix errors close to depth discontinuities.
  }
  \label{fig:dense_layers}
\end{figure*}

\Cref{fig:s2svss2d} illustrates the difference between our matching modes, and \cref{fig:pca} visualizes the structure learned by the tokens.
\Cref{fig:imc_clusters} shows the scene graphs recovered by our method on IMC2025~\citep{image-matching-challenge-2025}, and \cref{fig:dense_matches} compares our dense correspondences before and after refinement.
\Cref{fig:dense_layers} compares dense matches at different layers.
\Cref{fig:dg_samples} shows some examples of visual disambiguation, including some failure cases.
We finally compare our matches with LoMa~\citep{nordstrom2026loma} and RoMav2~\citep{edstedt2026romav2} in \cref{fig:match_examples}, including failure cases.

\newcommand{\dgimg}[1]{\includegraphics[height=1cm]{figures/dg_samples/#1}}
\newcommand{\dgrow}[3]{\resizebox{\linewidth}{!}{\dgimg{#1}\hspace{0.04cm}\dgimg{#2}\hspace{0.04cm}\dgimg{#3}}}
\newcommand{\dgrowfour}[4]{\resizebox{\linewidth}{!}{%
  \dgimg{#1}\hspace{0.04cm}\dgimg{#2}\hspace{0.04cm}\dgimg{#3}\hspace{0.04cm}\dgimg{#4}}}
\newcommand{\dgrowfive}[5]{\resizebox{\linewidth}{!}{%
  \dgimg{#1}\hspace{0.04cm}\dgimg{#2}\hspace{0.04cm}\dgimg{#3}\hspace{0.04cm}\dgimg{#4}\hspace{0.04cm}\dgimg{#5}}}
\begin{figure*}[t]
  \centering
  \dgrow{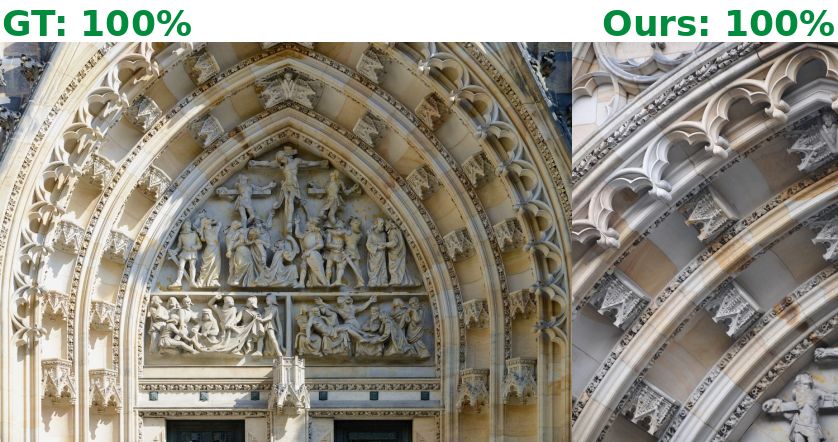}{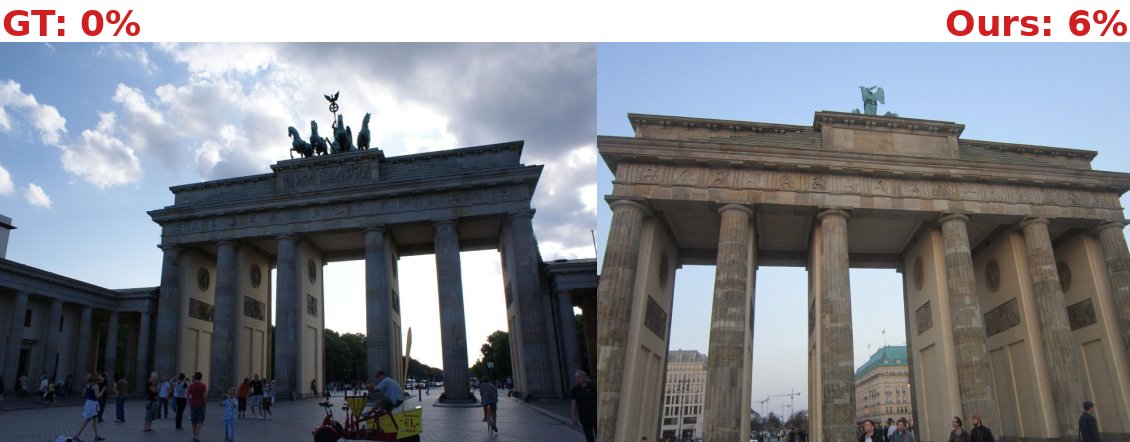}{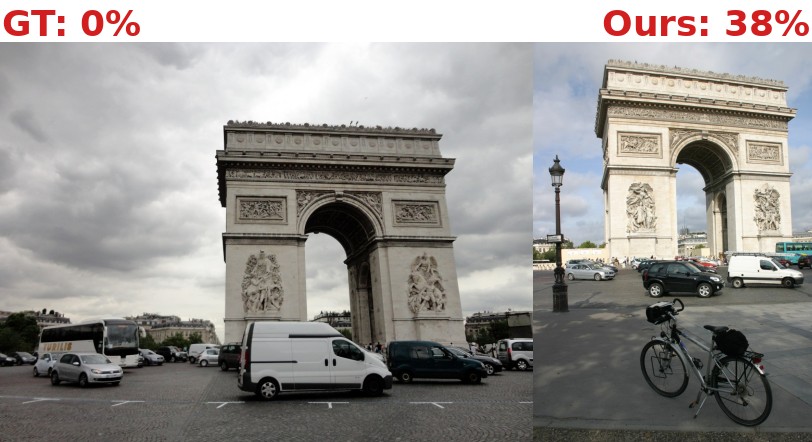}\\[2pt]
  \dgrowfive{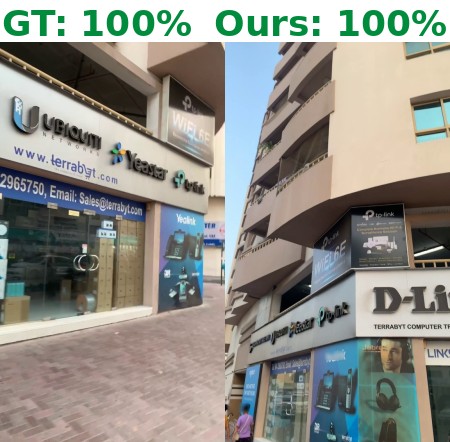}{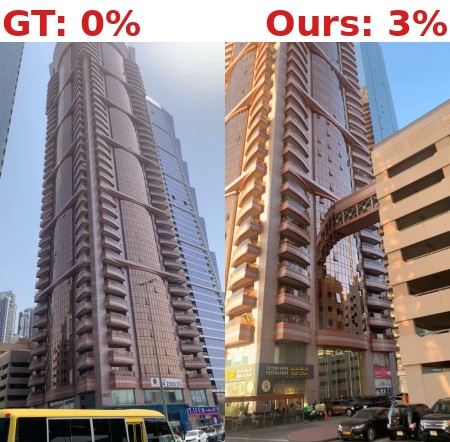}{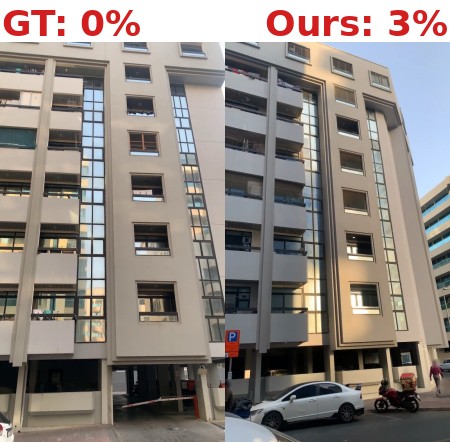}{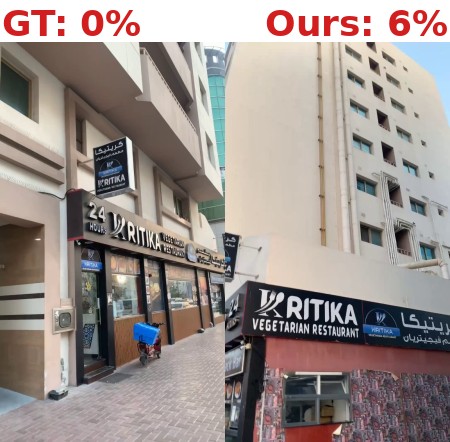}{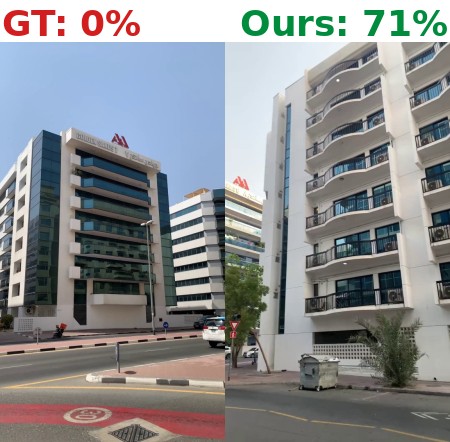}\\[2pt]
  \dgrowfour{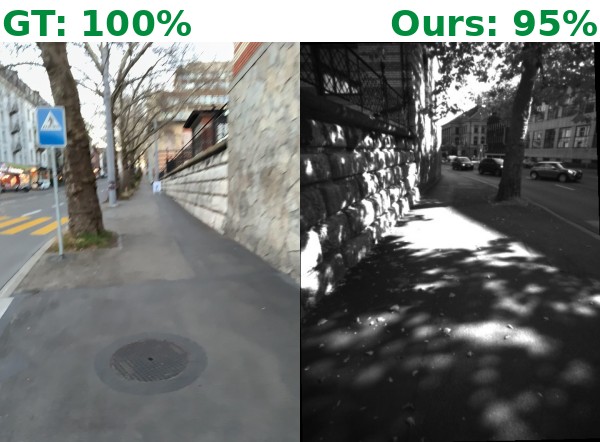}{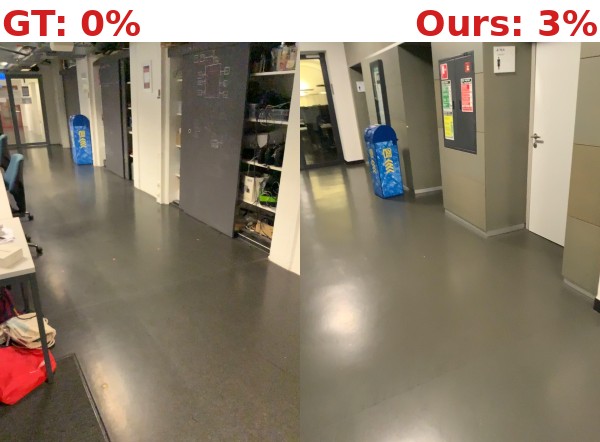}{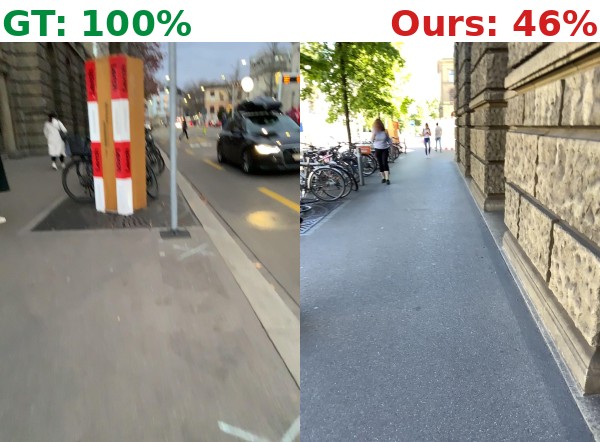}{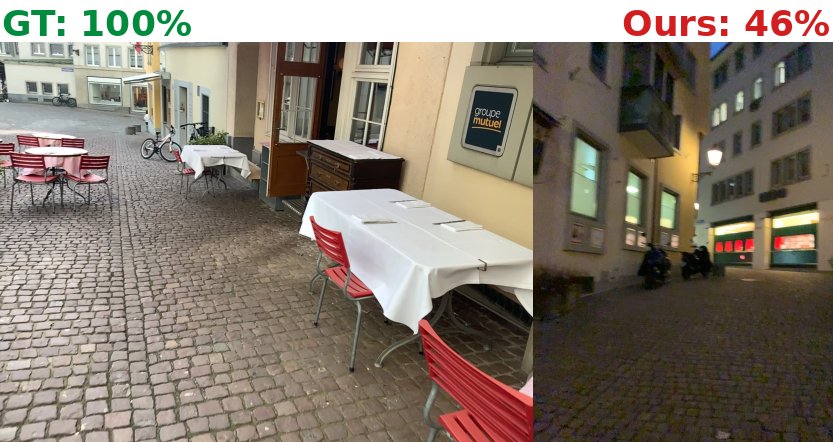}
  \caption{\textbf{Visual disambiguation examples.} We show some visual disambiguation examples from the test sets of DG~\citep{cai2023doppelgangers}, ViSym-F~\citep{xiangli2025doppelgangers++}, and LaMAR-DG~\citep{sarlin2022lamar}. Top left is the GT label, right our covisibility score. Our method effectively distinguishes hard covisible pairs, but sometimes mislabels under extreme appearance or viewpoint changes, or when context is missing (middle row failure, wrong facade).}
  \label{fig:dg_samples}
\end{figure*}

\newcommand{\matchrows}[2]{%
  \rotatebox[origin=c]{90}{\scriptsize #1}\hspace{2pt}%
  \begin{minipage}{\dimexpr\linewidth-12pt}
    \includegraphics[width=\linewidth]{figures/match_examples/#2_1.jpg}\\[1pt]
    \includegraphics[width=\linewidth]{figures/match_examples/#2_2.jpg}
  \end{minipage}\\[4pt]}
\newcommand{\matchheader}{%
  \hspace{12pt}%
  \begin{minipage}{\dimexpr\linewidth-12pt}
    \makebox[0.333\linewidth]{\small LoMa-G}\makebox[0.333\linewidth]{\small SP+RoMa v2}\makebox[0.333\linewidth]{\small SP+Ours}
  \end{minipage}\\[2pt]}

\begin{figure*}[p]
  \centering
  \matchheader
  \matchrows{MegaDepth}{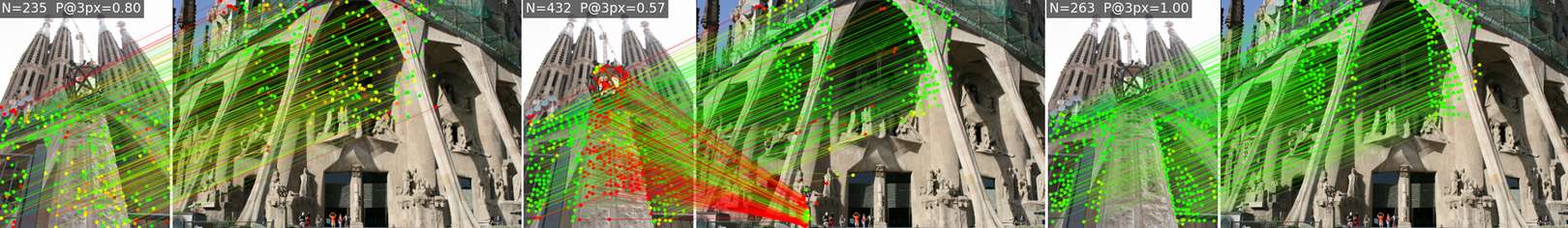}
  \matchrows{AerialMegaDepth}{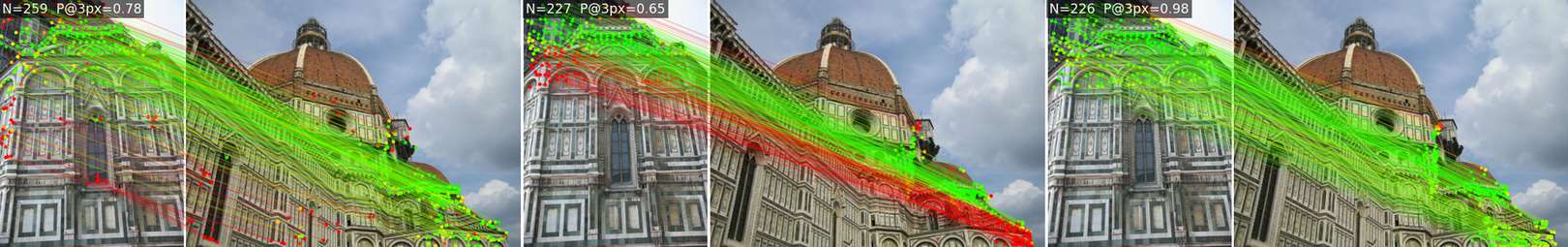}
  \matchrows{BlendedMVS}{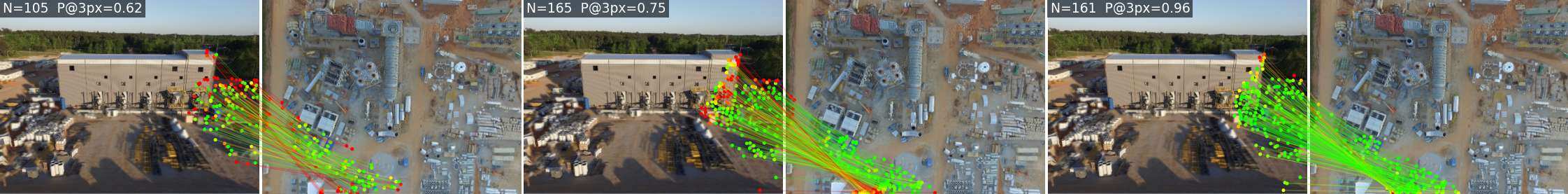}
  \matchrows{TerraSky3D}{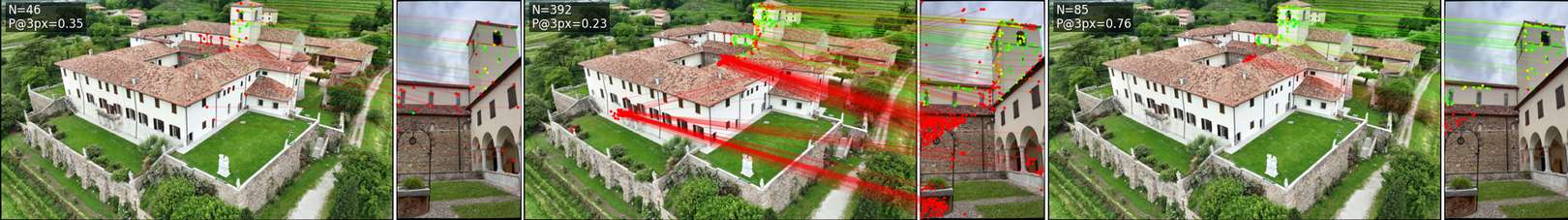}  %
  \caption{\textbf{Qualitative matches.} Correspondences of LoMa-G, SP+RoMav2, and SP+Ours on two pairs per dataset, colored by their reprojection error. We report the number of matches $N$ and the precision at 3\,px. LoMa gets 2048 keypoints, and RoMav2 and our method get 1024 keypoints for fair comparison.}
  \label{fig:match_examples}
\end{figure*}

\begin{figure*}[p]
  \ContinuedFloat
  \centering
  \matchheader
  \matchrows{ScanNet++}{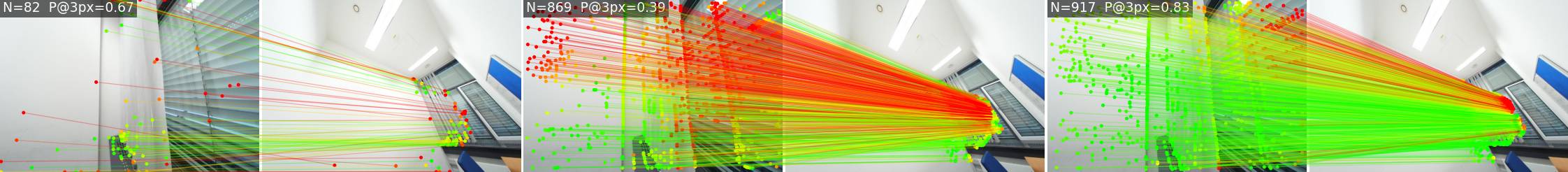}
  \matchrows{Hypersim}{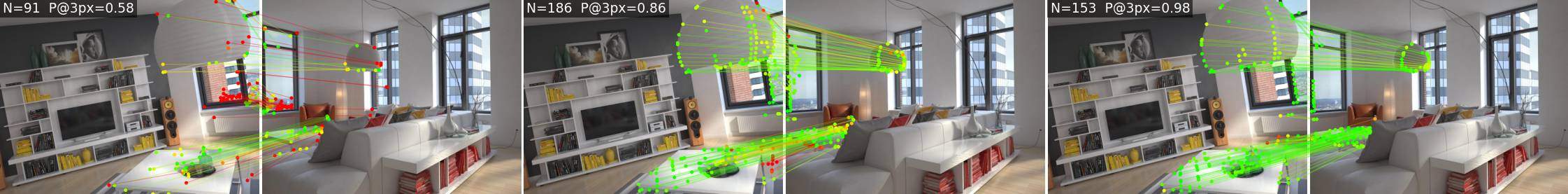}
  \matchrows{TartanAir}{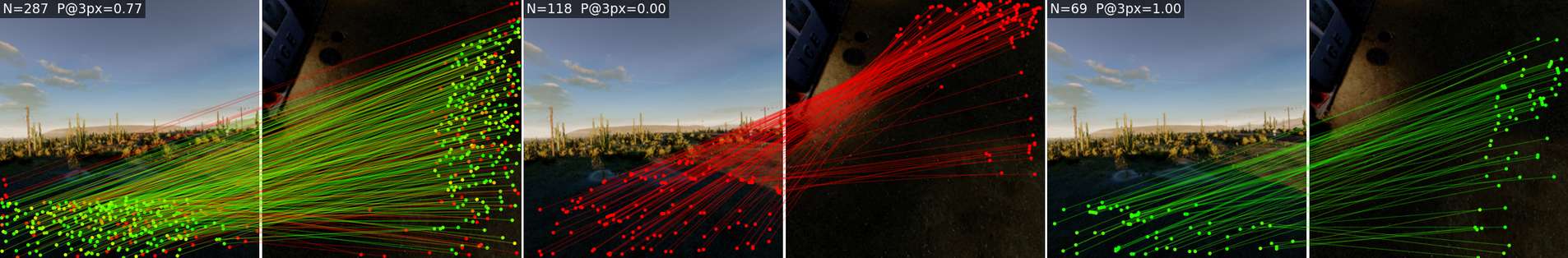}
  \rotatebox[origin=c]{90}{\scriptsize Failures}\hspace{2pt}%
  \begin{minipage}{\dimexpr\linewidth-12pt}
    \includegraphics[width=\linewidth]{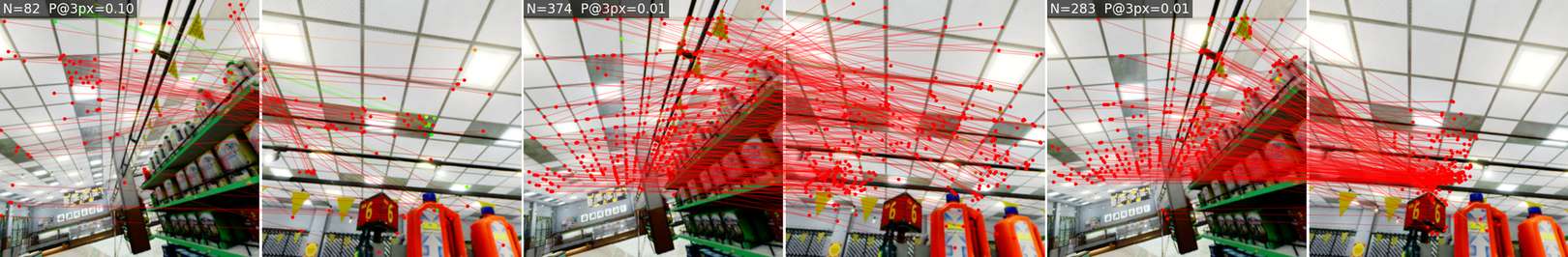}\\[1pt]
    \includegraphics[width=\linewidth]{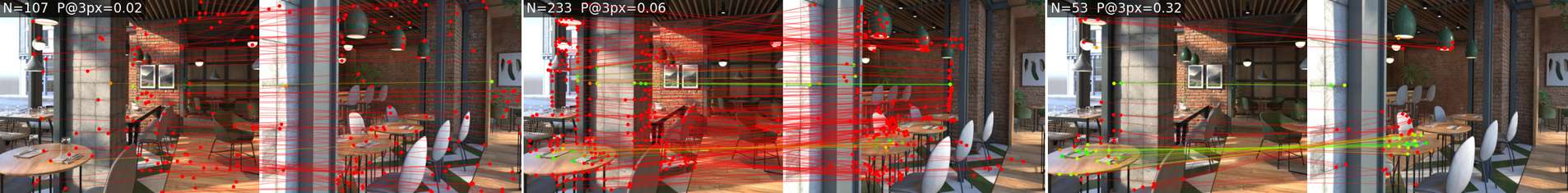}\\[1pt]
    \includegraphics[width=\linewidth]{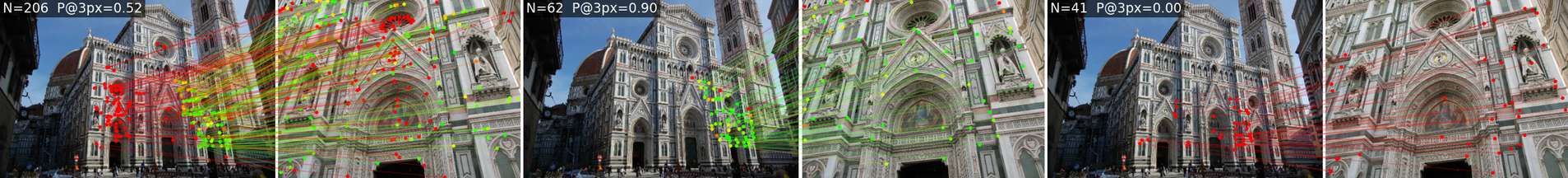}
  \end{minipage}
  \caption{\textbf{Qualitative matches (continued).} Bottom: failure cases on TartanAir, Hypersim, and MegaDepth. All methods fail under extreme rotation of a repetitive ceiling (top) and on a strongly self-similar scene (middle). On the repetitive facade (bottom), our method matches the wrong instance of a repeated element, while RoMav2 succeeds.}
\end{figure*}

\clearpage
\bibliography{main}
\bibliographystyle{iclr2027_conference}

\end{document}